\documentclass[letterpaper]{article} 
\usepackage{aaai2026}  
\nocopyright
\usepackage{times}  
\usepackage{helvet}  
\usepackage{courier}  
\usepackage[hyphens]{url}  
\usepackage{graphicx} 
\usepackage{natbib}  
\usepackage{caption} 
\usepackage{algorithm}
\usepackage{algorithmic}
\usepackage{xcolor}
\usepackage{amsmath}
\usepackage{multirow}
\usepackage{pdfpages}
\usepackage{newfloat}
\usepackage{listings}
\DeclareCaptionStyle{ruled}{labelfont=normalfont,labelsep=colon,strut=off} 
\floatstyle{ruled}
\newfloat{listing}{tb}{lst}{}
\floatname{listing}{Listing}
\title{Polymath: A Self-Optimizing Agent with Dynamic Hierarchical Workflow}

\author {
    Chia-Tung Ho\textsuperscript{\rm 1},
    Jing Gong\textsuperscript{\rm 1},
    Xufeng Yao\textsuperscript{\rm 2},
    Yunsheng Bai\textsuperscript{\rm 1},
    Abhishek B Akkur\textsuperscript{\rm 1},
    Haoxing Ren\textsuperscript{\rm 1}
}
\affiliations {
    \textsuperscript{\rm 1}Nvidia, Santa Clara, CA, USA\\
    \textsuperscript{\rm 2}Chinese University of Hong Kong, Hong Kong, China\\
    Corresponding Authors: chiatungh@nvidia.com,   haoxingr@nvidia.com
}

\usepackage{bibentry}

\begin{document}

\maketitle

\begin{abstract}
Large language models (LLMs) excel at solving complex tasks by executing agentic workflows composed of detailed instructions and structured operations. Yet, building general-purpose agents by manually embedding foundation models into agentic systems such as Chain-of-Thought, Self-Reflection, and ReACT through text interfaces limits scalability and efficiency.
Recently, many researchers have sought to automate the generation and optimization of these workflows through code-based representations. However, existing methods often rely on labeled datasets to train and optimize workflows, making them ineffective and inflexible for solving real-world, dynamic problems where labeled data is unavailable.
To address this challenge, we introduce Polymath, a self-optimizing agent with dynamic hierarchical workflow that leverages the flexibility of task flow graphs and the expressiveness of code-represented workflows to solve a wide range of real-world, dynamic problems.
The proposed optimization methodology integrates multi-grid-inspired graph optimization with a self-reflection-guided evolutionary algorithm to refine workflows without labeled data.
Experimental results on six benchmark datasets across coding, math, and multi-turn QA tasks show that Polymath achieves 8.1\% average improvement over state-of-the-art baselines. 
We will make the source code publicly available upon acceptance.
\end{abstract}

\section{Introduction}
Large Language Models (LLMs)~\cite{GPT4, anthropic2024claude35} have demonstrated remarkable capabilities across a wide range of domains, from code generation and data analysis to decision-making and complex reasoning. 
However, to solve complex real-world problems, their effectiveness often hinges not just on the model itself but on carefully crafted agentic workflows-structured sequences of prompts, tool interactions, and logic designed by humans, such as chain-of-thought (CoT) planning and reasoning~\cite{wei2022chain, hu2023thought}, 
ReACT and tool use~\cite{yao2022react, schick2023toolformer}, and self-reflection~\cite{shinn2023reflexion, madaan2023self}.
While these agentic workflows enable LLMs to solve challenging problems, they are typically hand-engineered, task-specific, and labor-intensive to design and maintain. 
As the demand for LLM-driven applications expands, this reliance on manual workflow construction becomes a bottleneck. It limits the scalability of LLM systems, slowing adaptation to new domains, and hindering the transfer of skills across tasks~\cite{tang2023verifai}. 
Therefore, automating agentic workflows for solving versatile and diverse tasks has emerged as a critical need. 

Many recent works focus on automating agentic workflow discovery to reduce human involvement~\cite{khattab2024dspy, yuksekgonul2024textgrad, liu2023dynamic, hu2024automated}, yet full automation remains unsolved. DSPy~\cite{khattab2024dspy} requires manual setup, while methods like TextGrad~\cite{yuksekgonul2024textgrad} and GPTSwarm~\cite{zhuge2024gptswarm} struggle to capture the diversity of workflows needed for broad task generalization~\cite{yu2023thought, yang2024buffer, sun2023indeterminacy}, since their optimization objectives cannot represent the breadth of potential workflows. 
Although ADAS~\cite{hu2024automated} and AFlow~\cite{zhang2024aflow} improve expressiveness by representing workflows as code and refining them via execution feedback, they rely heavily on existing validation data and aim to generalize across task categories, limiting their adaptability to dynamic, real-world problems and task-specific challenges.
On the other hand, Data Interpreter~\cite{hong2024data} proposed a task graph on top of a programmable node flow, but the approach lacks efficient self-learning and optimization.
This highlights the critical need for more effective and adaptive techniques to fully automate the workflow generation for dynamic, real-world problems to accelerate the application of LLMs across domains.

In this work, we propose Polymath, a self-optimizing agent featuring a dynamic hierarchical workflow that leverages flexible task flow graphs combined with expressive, code-based workflows to tackle a broad range of real-world, dynamic problems. 
Moreover, we propose a novel hierarchical workflow optimization methodology, from multi-grid-inspired task flow graph optimization to an online self-reflection-guided evolutionary algorithm for code-represented workflow enhancement through LLM-based evaluators without the need for labeled datasets. Our contributions are as follows.

\begin{figure*} [!t]
	\centering
    \includegraphics[width=1.85\columnwidth]{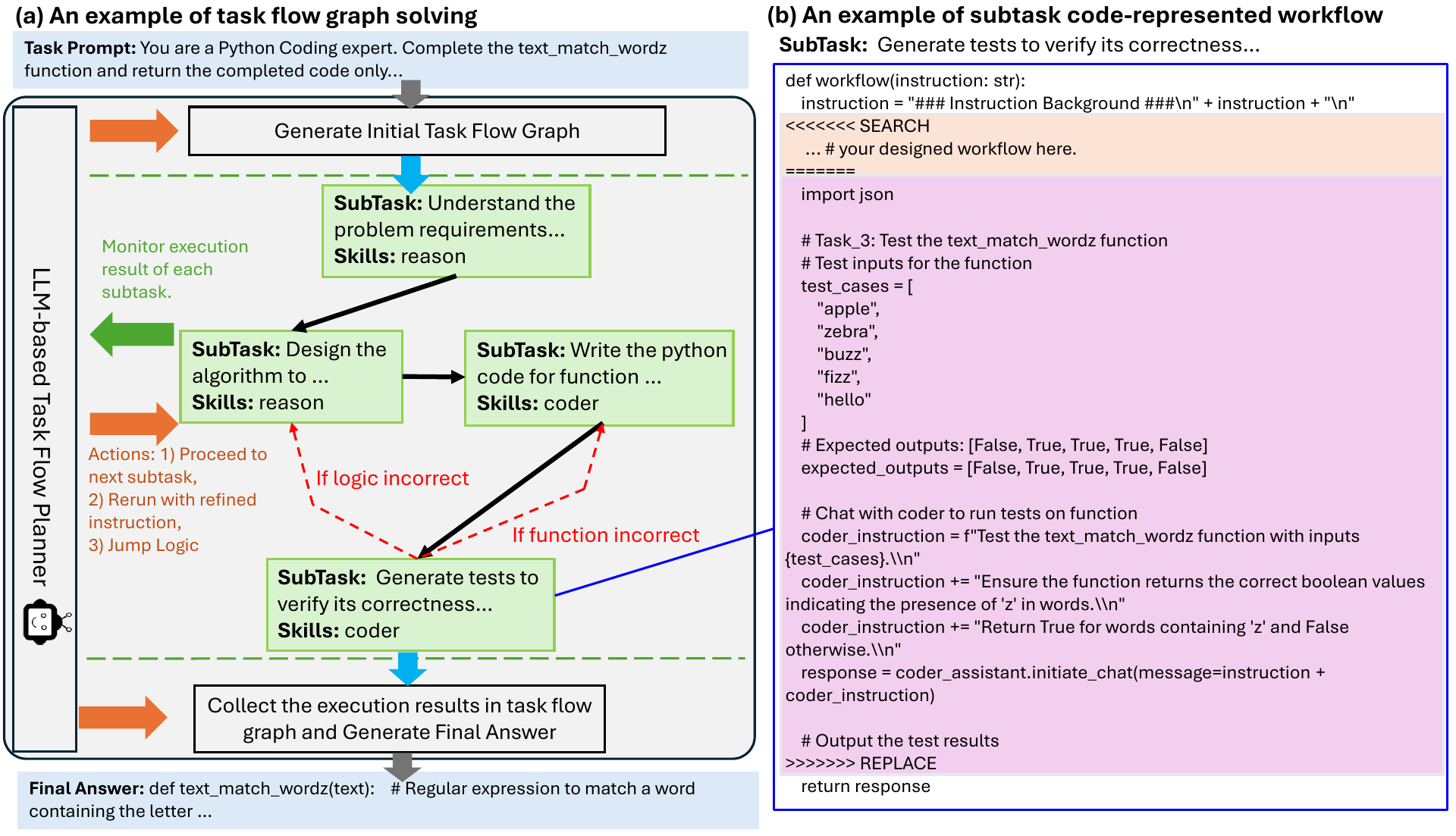}
	\vspace{-0.3cm}
    \caption{An illustration of (a) task flow graph solving and (b) code-represented subtask workflow.}
	\label{TaskGraphSubtaskExampleFig}
    \vspace{-0.4cm}
\end{figure*}

\begin{itemize}
    \item We propose a self-optimizing agent with dynamic hierarchical workflow that leverages the flexibility of task flow graphs and the expressiveness of subtask-level code-represented workflows to solve a wide range of real-world, dynamic problems. The task flow graph employs a divide-and-conquer approach to decompose and execute subtasks based on the topological order, while the code-represented workflow ensures stable and robust execution without hallucinations.   
    \item We develop a novel hierarchical workflow optimization that combine multi-grid-inspired task flow graph optimization with a self-reflection-guided evolutionary algorithm to enhance code-represented workflows on the fly using feedback from reasoning LLMs~\cite{jaech2024openai}, eliminating the need for labeled datasets.
    \item We conduct extensive and holistic studies of text-represented workflows, code-represented workflows, and the proposed hierarchical workflow on HumanEval, MBPP, MATH$_{{lv5*}}$, GSM8K, HotpotQA, and DROP datasets. We demonstrate that Polymath achieves 8.1\% better average scores over state-of-the-art baselines.
    \item We perform studies on the real-world industrial case in the hardware design area that requires to digest multiple files, block diagrams, and an approximately 100-page datasheet to demonstrate the capability of the proposed Polymath to solve real-world problem.
\end{itemize}
The remaining sections are organized as follows. \textcolor{black}{We first review related works on Agentic workflows and workflow optimization.}
Then, we introduce and describe our novel workflow generation methodology in details. Lastly, we present main \textcolor{black}{experimental results} and conclude the paper.

\section{Related Work}
{\bf{Agentic System}}: Researchers have developed various building blocks and design patterns for agentic system across diverse applications, such as chain-of-thought (CoT) planning and reasoning~\cite{wei2022chain, hu2023thought}, self-consistency~\cite{wang2022self}, memory structures~\cite{lewis2020retrieval, zhang2024survey}, 
ReACT and tool use~\cite{yao2022react, schick2023toolformer}, self-reflection~\cite{shinn2023reflexion, madaan2023self}, and graph-based planning~\cite{yao2023tree, besta2024graph, ho2025verilogcoder}.
Agentic system methodologies can be broadly categorized into general and domain-specific types.
General agentic methodologies focus on universal problem-solving~\cite{wang2024openhands, wang2023unleashing}, while domain-specific agentic approaches aim to build effective processes for solving particular types of problems, such as software coding~\cite{yang2024swe, huang2023agentcoder, xia2024agentless, sohrabizadehnemotron, liu2025sew, aider}, mathematics~\cite{zhong2024achieving}, hardware design~\cite{nainani2025timing, ho2025verilogcoder, chang2025drc, lai2025analogcoder}, and log parsing~\cite{liu2022uniparser, le2023log}.
Although these agentic approaches are effective, their workflows often rely on manual fine-tuning and development, which makes it challenging to cover the wide variety of tasks across different application domains.
Therefore, developing effective and efficient automated workflow generation and optimization is both essential and critical.

\noindent{\bf{Agentic System Optimization}}:
Recent works on automatic agentic system optimization focus on three domains: prompt optimization, hyperparameter optimization, and agentic workflow optimization. 
Existing prompt optimization methods leverage a fixed agentic workflow to optimize prompts~\cite{fernando2023promptbreeder, yuksekgonul2024textgrad, yang2023large, khattab2024dspy}. Hyperparameter optimization work~\cite{saad2024archon} focuses on tuning predefined parameters. 
These approaches require moderate human effort for task-specific design and are limited in their ability to automatically optimize for new tasks.

To address the challenge, automated agentic workflow optimization aims to optimize entire workflow structures for fully automated generation.
For example, \cite{li2024autoflow, zhou2024symbolic} optimize workflow sequences through text representations, while GPTSwarm~\cite{zhuge2024gptswarm} uses graph-represented workflows with reinforcement learning.
However, both approaches struggle to represent workflows with conditional states and complex looped task flows due to limitations in text and graph expressiveness.
Recently, ADAS~\cite{hu2024automated}, AFlow~\cite{zhang2024aflow}, and EvoFlow~\cite{zhang2025evoflow} have improved expressiveness by representing workflows as code and refining them via execution feedback.
Nevertheless, they rely heavily on existing validation data and focus on generalizing across task categories, limiting adaptability to dynamic real-world problems and task-specific challenges. 
Additionally, AFlow optimized a workflow for a entire task category, which is inefficient or ineffective for solving different levels of problems within the same task category.

The proposed self-optimizing and dynamic hierarchical workflow generation methodology leverages flexible divide-and-conquer task flow graphs on top of expressive code-represented workflows to solve a wide range of dynamic, real-world problems. 
Furthermore, the novel hierarchical optimization technique orchestrates optimization across both graph structure and code through a multi-grid-inspired task flow graph optimization and a self-reflection-guided evolutionary algorithm, enhancing code-represented workflows on the fly without requiring labeled datasets.

\begin{figure*} [!t]
	\centering
    \includegraphics[width=1.9\columnwidth]{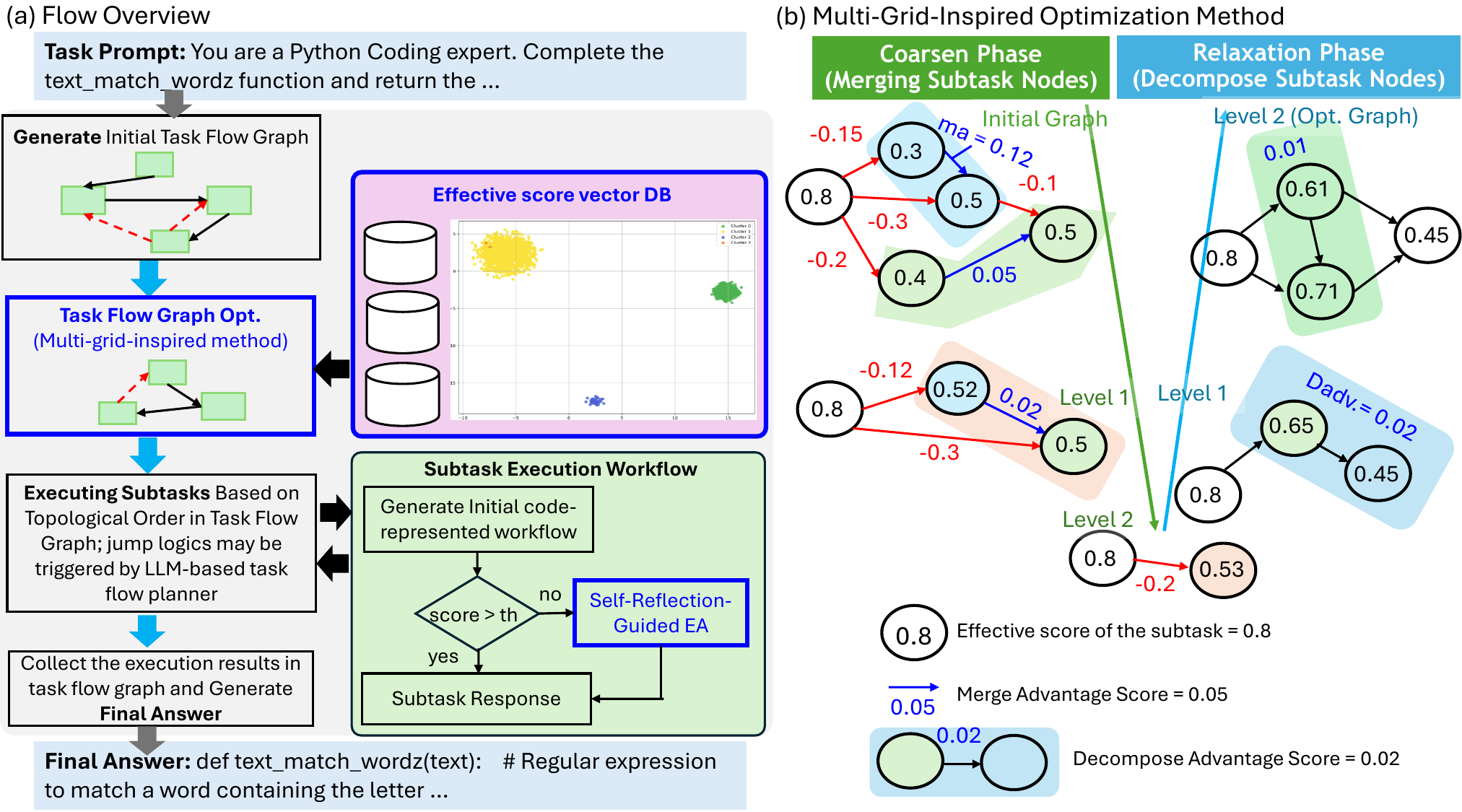}
    \vspace{-0.2cm}
    \caption{(a) Flow overview of self-optimizing agent with dynamic hierarchical workflow. The blue boxes are the key components of self-optimization. The black boxes are introduced in the preliminary section. (b) An illustration of multi-grid-inspired task flow graph optimization.}
	\label{FlowOverviewFig}
    \vspace{-0.4cm}
\end{figure*}

\section{Preliminary} \label{PrelimSection}
In this section, we first introduce the core components of the proposed task flow graph and code-represented workflows, followed by the formulation of the dynamic hierarchical workflow generation problem.

\subsection{Task Flow Graph}
Planning is a core module for an agent~\cite{wang2024survey, weng2023agent} to decompose complex tasks into manageable subtasks.
Earlier works such as CoT~\cite{wei2022chain, hu2023thought} decompose complex tasks sequentially, but this sequential execution is often insufficient for addressing more complex problems. 
Generating a task flow graph and executing tasks node-by-node has shown promising results in breaking down complex tasks into manageable subtasks~\cite{besta2024graph, ho2025verilogcoder}.
Inspired by prior works, We leverage the task flow graph, $G(T, E)$, for planning, which initially decomposes a problem into manageable subtasks (see Figure~\ref{TaskGraphSubtaskExampleFig}(a)). 
The execution of the task flow graph, $G$, is monitored by a LLM-based task flow planner, whose actions include: 1) proceeding to the next subtask, 2) rerunning a subtask, and 3) applying jump logic.
Finally, the task flow planner produces the final answer based on all the execution results from all subtasks.

In $G(T, E)$, the nodes, $T=(t_1, t_2, ..., t_n)$, represent subtasks, and the edges, $E=(e_1, e_2, ..., e_m)$, represent the task dependencies or jump logic relations.
We leverage the reasoning ability of LLMs to decompose a task-oriented input, $x$, into a set of subtasks $T$ that can be individually solved and verified. 
Each subtask $t_i$ receives input $r_i$, which encodes intermediate results from subtasks on which $t_i$ depends, as defined by $G$.
We can write the solving process as graph $G$ that embodies the entire subtasks:
\begin{equation}
\hat{y} = G\left( \left\{ t_i(x, r_i) \right\}_{i=1}^{n}, E \right)
\end{equation}

\noindent{\bf{Task Flow Graph Problem Formulation}}:
The primary challenge lies in determining the appropriate complexity, size, and relationships of each subtask.
Improving the effective score $s$ of each subtask (i.e., efficiency and completeness) and edge relationships involves achieving an optimal task flow graph. 
As a result, We can formulate this as the following optimization problem:
\begin{equation}
\begin{aligned}
\mathbf{t}^\star &= \arg\max_{\mathbf{t} \in \mathcal{T}(x, E)} \frac{1}{|T|} \sum_{i=1}^{n} s\big(t_i(x, r_i)\big), \\
G^* &= \arg\max_{G} \; F\left( G\left( \mathbf{t}^\star, E \right), x\right) 
\end{aligned}
\label{TaskFlowGraphEq}
\end{equation}
Where $t=(t_1, ... ,t_n)$ is drawn from the feasible set $\mathcal{T}(x, E)$, the set of feasible subtask configurations given input $x$, and graph dependencies $E$. $F$ is the evaluation function for the given input $x$.

\subsection{Code-Represented Subtask Workflow}
Each subtask in task flow graph is represented as a code-based workflow that takes input instructions derived from the decomposed task flow graph and outputs response, as illustrated in Figure~\ref{TaskGraphSubtaskExampleFig}(b). 
The assistant, assistant instruction prompt, and execution flow are generated within a Replace block for the "Generate tests to verify its correctness" subtask in the task flow graph (see Figure~\ref{TaskGraphSubtaskExampleFig}(b)).

\noindent{\bf{Search Space}}: 
The code-represented subtask workflow $W$ is assembled by combining a series of LLM assistant invoking nodes $V$. 
The key parameters of $W$ are as follows.
\begin{itemize}
    \item {\bf LLM Assistants} $A$: The specific LLM assistant at $v_i$. 
    Each assistant is a functional unit capable of performing a complete task, such as coding, reasoning, or file reading. For example, a coding assistant not only generates code but also executes it and returns the result.
    \item {\bf Assistant Instruction Prompt} $P_i$: The input instructions or task descriptions provided to each node $v_i$.
    \item {\bf Links} $L$: The abstract structures define node relationships and govern execution flow.
\end{itemize}

The search space $S_w$ for a workflow optimization problem encompasses all possible configurations of node parameters and link structures.
\begin{equation}
S_w = \{\, (V(A, P), L) \mid  A \in \mathcal{A}, P \in \mathcal{P}, L \in \mathcal{L} \,\}
\end{equation}
\noindent Where $\mathcal{A}$, $\mathcal{P}$, $\mathcal{L}$ represent the sets of possible LLM assistants, assistant instruction prompts, and links, respectively.

\noindent{\bf{Subtask Workflow Problem Formulation}}:
Given a task flow graph status $K$, a subtask $t$, and an evaluation function $u$,
the goal of subtask workflow optimization is to find a workflow $W$ that maximizes $u(W, K, t)$.
We formulate the subtask workflow problem as below.
\begin{equation}
W^* = \arg\max_{W \in S_w} u(W, K, t)
\label{WorkflowObjEq}
\end{equation}
\noindent where $W^*$ is the optimal workflow that maximizes the evaluation function $u$ for the given task flow graph status $K$ and subtask $t$.

Compared to recent works~\cite{hu2024automated, zhang2024aflow, zhang2025evoflow} and AFlow~\cite{zhang2024aflow} using code to represent entire workflow, the proposed hierarchical workflow with task flow graph on top of code-represented subtask workflow can react to the unexpected error execution of code-represented workflow dynamically and refine task flow graph to solve the task.


\section{Optimization Methodology}
We propose a novel hierarchical optimization methodology that orchestrates optimization across both the task flow graph structure and the subtask coding workflows.
Our approach integrates a multi-grid-inspired task flow graph optimization with a self-reflection-guided evolutionary algorithm (EA) to dynamically enhance code-represented workflows without requiring labeled training or validation datasets. 
Figure~\ref{FlowOverviewFig}(a) shows the overall workflow generation process, including on-the-fly hierarchical optimization.
After constructing the initial task flow graph, we perform task flow graph optimization based on an effective score vector database.
For each subtask, the initial workflow is evaluated by LLM judge; if the score is smaller than a predefined threshold, the self-reflection-guided EA is applied to further optimize the code-represented workflow.
Upon completing all the tasks in the task flow graph, the task flow planner aggregates the execution results of all subtasks to produce the final answer. 
See the appendix for more detailed studies.

\subsection{Task Flow Graph Optimization}
The goal of task flow graph optimization is to balance the complexity and success rate of subtasks and their relationships in order to find $G^*$, as defined in Eq. (\ref{TaskFlowGraphEq}).
We first introduce the concept of the subtask effective score, followed by the proposed multi-grid-inspired task flow graph optimization methodology.

\noindent{\bf{Subtask Effective Score}}:
The effective score, denoted as $s$, is computed as the product of the workflow complexity score and the completeness score, thereby capturing a trade-off between task complexity and completeness.
For brevity, we use $s_i$, $d_i$, and $c_i$ to represent the effective score, complexity score, and completeness score of subtask $t_i$, respectively. Here, we construct a effective score vector database, $D_s$, by using LLM judges to evaluate $c_i$ and $d_i$, and generate corresponding reflections of each subtask based on its task content, workflow, and response in history.
In addition, we cluster historical subtasks based on their task content vectors and compute the statistical distributions of $d_i$ and $c_i$ within each cluster to mitigate noise in LLM-based evaluations.
To estimate the effective score $\hat{s}_k$ of a new subtask $t_k$, we first retrieve the top-$K$ most similar tasks from $D_s$, then aggregate the complexity and completeness statistics from their clusters. 
A LLM operator $\epsilon$ uses these statistics and $t_k$ to estimate $\hat{d}_k$ and $\hat{c}_k$, from which $\hat{s}_k$ is computed as below.

\begin{equation}
\begin{aligned}
\hat{d}_k, \hat{c}_k &= \epsilon\big( t_k, \text{Top-}K(t_k, D_s),\, \mu_c^{(k)}, \sigma_c^{(k)},\, \mu_d^{(k)}, \sigma_d^{(k)} \big), \\
\hat{s}_k &= \hat{d}_k \cdot \hat{c}_k
\end{aligned}
\label{RetrieveEq}
\end{equation}

\noindent{\bf{Multi-Grid-Inspired Graph Optimization}}: 
Finding $G^*$ requires exploring an enormous solution space of subtasks and their relationships, which poses significant computational challenges. 
For instance, given an initial graph $G_0(T_0, E_0)$, the complexity of enumerating all possible merging combinations is $O(2^{|T_0|})$. 
To improve the efficiency of optimizing $G_0$, we propose a novel multi-grid-inspired optimization method~\cite{trottenberg2001multigrid, karypis1999multilevel}. 
This approach applies a typical V-cycle procedure to iteratively coarsen and relax the task flow graph, leveraging the advantage effective scores in $G_0$ (see Figure~\ref{FlowOverviewFig}(b)).


\begin{algorithm}[!t]
\caption{Task Flow Graph Coarsen Algorithm}
\label{CoarsenAlg}
\begin{algorithmic}[1]
\REQUIRE Edge list $E = \{(i, j, ma_{i,j})\}$ sorted in descending order of $ma_{i, j}$
\ENSURE Selected merge pairs $\mathcal{M}$
\STATE $\mathcal{M} \gets \emptyset$
\STATE $\text{used} \gets \emptyset$
\FORALL{$(i, j, ma_{i, j}) \in E$}
    \IF{$ma_{i, j} < 0$}
        \STATE \textbf{break}
    \ENDIF
    \IF{$u \notin \text{used}$ \AND $v \notin \text{used}$}
        \STATE $\mathcal{M} \gets \mathcal{M} \cup \{(i, j, ma_{i,j})\}$
        \STATE $\text{used} \gets \text{used} \cup \{i, j\}$
    \ENDIF
\ENDFOR
\RETURN $\mathcal{M}$
\end{algorithmic}
\end{algorithm}

\noindent{\textit{\underline{Coarsen Phase}}}:
Coarsen phase aims to merge the adjacent subtask nodes that exhibit a positive merge advantage score, $ma$. 
Given a task flow graph $G_0$, the estimated effective score of each subtask node is computed using Eq. (\ref{RetrieveEq}). 
For each edge connects $t_i$ and $t_j$ in $G_0$, we estimate the effective score $\hat{s}_{i,j}$ of merging $t_i$ and $t_j$ by substituting $k$ with $(i,j)$ in Eq. (\ref{RetrieveEq}).  
The $ma_{i,j}$ is then calculated as:
\begin{equation}
ma_{i,j} = \hat{s}_{i,j} - \frac{\hat{s}_i + \hat{s}_j}{2}
\label{MergeAdvEq}
\end{equation}

We maximize total merge advantage, with each node merged at most once per coarsening level (Eq. (\ref{MergeProblemFormulationEq})).
$w_{u,v}$ indicates whether nodes $u$ and $v$ are merged.
\begin{equation}
\begin{aligned}
\max_{x} & \sum_{(u,v) \in E} ma_{u,v} \cdot w_{u,v} \\
\text{s.t.} \sum_{(u,v) \in E : u = i || v = i} & w_{u,v} \leq 1, \; w_{u,v} \in \{0, 1\}, \; \forall i \in V 
\end{aligned}
\label{MergeProblemFormulationEq}
\end{equation}

We adopt a greedy approximation algorithm (Algorithm~\ref{CoarsenAlg}) to solve Eq. (\ref{MergeProblemFormulationEq}) at each coarsening level for efficiency. 
First, all edges are sorted in descending order of $ma_{i,j}$.
Then, for each edge, the corresponding node pair is selected for merging if neither node has been previously merged in this level (Lines 7–9). 
Finally, we return the merged node pairs and project the task flow graph to next coarser level. 
This coarsening procedure continues until either the predefined coarsening level is reached or all $ma_{i,j}$ values become negative.

\noindent{\textit{\underline{Relaxation Phase}}}:
The relaxation phase focuses on decomposing complex subtasks to improve the average effective scores.  
Algorithm~\ref{RelaxAlg} shows the relaxation methodology at each relaxation level.
First, we iterate over each subtask node in the \textcolor{black}{graph obtained from the coarsen phase} and leverage an LLM to decompose the subtask (Lines 1–2).
After obtaining a valid subgraph $G_s$, we compute the decompose advantage, $da_t$, of the subtask node $t$ (Line 6).
If $da_t$ is positive, we replace the original subtask node $t$ with the subgraph $G_s$ (Lines 7–12).
Finally, the algorithm returns the relaxed graph $G'$ (Line 14).
This relaxation procedure continues until either the predefined relaxation level is reached or all $da_t$ values become negative.

\begin{algorithm} [!t]
\caption{Task Flow Graph Relaxation Algorithm}
\label{RelaxAlg}
\begin{algorithmic}[1]
\REQUIRE Graph $G = (T, E)$ with node effective scores $s_t$, decomposition operator $h(t, l=4)$
\ENSURE Updated graph $G'$ after relaxation

\FORALL{node $t \in T$}
    \STATE Query $G_s = (T_s, E_s) \gets h(t, l)$ \quad \COMMENT{LLM proposes candidate subgraph with max 4 nodes limitation}
    \IF{$T_s = \emptyset$}
        \STATE \textbf{continue} \quad \COMMENT{No decomposition proposed}
    \ENDIF
    \STATE Compute $\text{da}_t \gets (\frac{1}{|T_s|} \sum_{u \in T_s} \hat{s}_u) - \hat{s}_t$
    \IF{$\text{da}_t > 0$}
        \STATE Remove node $t$ and its incident edges from $G$
        \STATE Insert nodes $T_s$ and edges $E_s$ into $G$
        \STATE Connect incident edges $(*, t)$ to $G_s$'s root nodes
        \STATE Redirect outgoing edges $(t ,*)$ to $G_s$'s terminals
    \ENDIF
\ENDFOR
\RETURN Relaxed graph $G'$
\end{algorithmic}
\end{algorithm}
\vspace{-0.2cm}

\subsection{Subtask Workflow Optimization}
The goal of subtask workflow optimization is to maximize the evaluation score for a given task flow graph state $K$ and subtask $t$.
Inspired by recent works~\cite{novikov2025alphaevolve, openevolve} that demonstrate promising capabilities of leveraging LLMs for code optimization, we propose a self-reflection-guided EA. 
This approach enhances code optimization by incorporating additional textual gradients derived from self-reflections and is built on top of the OpenEvolve framework~\cite{openevolve}.


\noindent{\bf{Prompt Sampler and Code Generation}}:
The prompt sampler aggregates multiple previously discovered workflows sampled from the database, along with their scores, self-reflection guidance, and problem descriptions.
Then, we leverage LLMs to generate the next-generation workflows which follows the gradient of scores and textual gradient.

\begin{figure*} [!t]
	\centering
    \includegraphics[width=1.875\columnwidth]{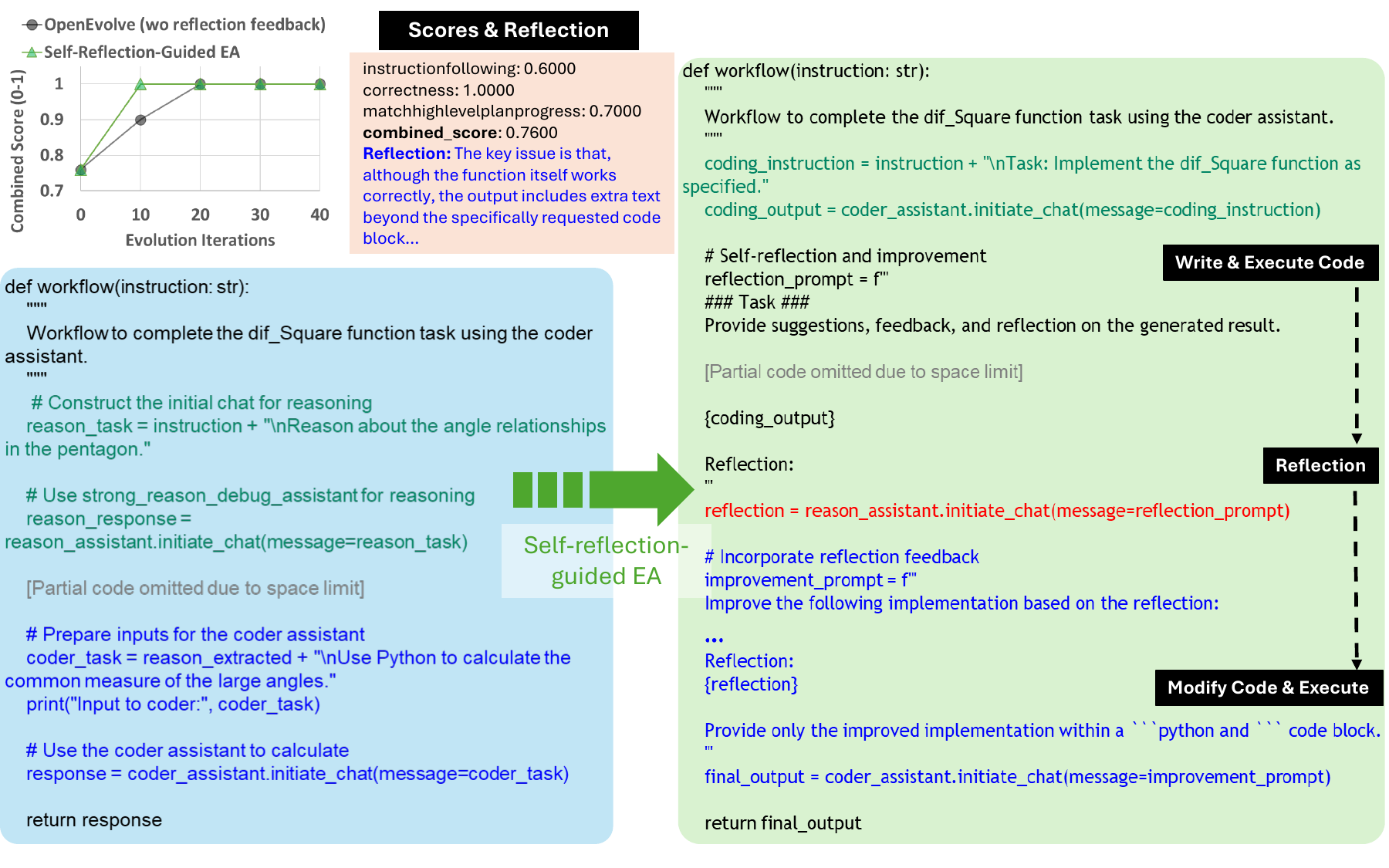}
    \vspace{-0.4cm}
    \caption{An example of self-reflection-guided evolutionary algorithm.}
	\label{EAEvaluationFig}
    \vspace{-0.4cm}
\end{figure*}

\noindent{\bf{Evaluation}}: 
Every new workflows generated are automatically evaluated for tracking evolutionary progress and selecting which ideas to propagate in future generations.
We develop the evaluation function (i.e., $u$ in Eq. \ref{WorkflowObjEq}) that is aware of the task flow graph status and provides the multi-objective scores along with self-reflections generated by LLM judges.
The multi-objective scores of $u$ are as follows: \textbf{InstructionFollowing score} measures how well the task output follows the user instructions; \textbf{Correctness score} evaluates the accuracy of the task output relative to the user request; \textbf{MatchHighLevelPlanProgress score} assesses how well the current subtask output aligns with the expectations defined in the task flow graph; and \textbf{Combined score} represents a weighted sum of the above scores.

Figure~\ref{EAEvaluationFig} illustrates an example of multi-objective scores, self-reflections, and the evolved code after applying the self-reflection-guided EA. 
Notably, the self-reflection-guided EA achieves the perfect combined score within 10 evolution iterations, whereas OpenEvolve requires 20 iterations to reach the same combined score.


\noindent{\bf{Evolution}}:
The evolution process continually generates a growing pool of candidate solutions, each annotated with evaluation results (scores, self-reflections, and program outputs), which are stored in a program database.
A key challenge is to balance exploration and exploitation for continuously improving the best programs while maintaining diversity to encourage exploration of the entire search space.
To address this, we leverage the OpenEvolve framework, which integrates MAP-Elites~\cite{mouret2015illuminating}, island-based population models~\cite{romera2024mathematical, tanese1989distributed}, and exploratory program sampling strategies to effectively maintain this balance.

\section{Experiments}
Our implementation is developed in Python and the self-reflection-guided EA is built on top of OpenEvolve~\cite{openevolve}. 
In all experiments, we use consistent prompts and settings within Polymath agentic flow to demonstrate the self-optimizing capability. 
We construct the effective score vector database using unoptimized runs from the HumanEval and MATH benchmarks for the multi-grid-inspired task flow graph optimization.
Each subtask uses a 0.8 threshold for self-reflection-guided EA, starting with an empty program database and up to 15 iterations.

\noindent{\bf{Dataset}}: We evaluate our approach on six public benchmarks: 
(1) \textit{Multi-turn QA:} HotpotQA~\cite{yang2018hotpotqa} and DROP~\cite{dua2019drop};
(2) \textit{Coding:} HumanEval~\cite{chen2021evaluating} and MBPP~\cite{austin2021program};
(3) \textit{Math:} MATH~\cite{hendrycks2021measuring} and GSM8K~\cite{cobbe2021training}.
Following~\cite{zhang2024aflow}, we use the full datasets for GSM8K, HumanEval, and MBPP, sample 1,000 examples from HotpotQA and DROP, and select the high-difficulty subset (difficulty level 5) for MATH$_{{lv5*}}$~\cite{hong2024data}. 
We run experiments on these six benchmarks without separating them into validation and test sets.  
Additionally, we include an industrial case study in hardware design, which involves digesting multiple files, a block diagram, and an approximately 100-page datasheet, to demonstrate Polymath’s capability in solving real-world problems.

\noindent{\bf{LLM and Assistant Settings}}: 
We use GPT-4o-1120~\cite{openai2024gpt4o} as the core model in Polymath, supported by a set of assistants: a coder assistant (GPT-4o-1120), a reasoning assistant (o1-1217~\cite{openai2024o1}), an image reader (GPT-4o-1120), and a file reader (GPT-4o-1120).  
We implement the coder assistant, image reader and file reader using the Autogen framework~\cite{wu2023autogen}.

\noindent{\bf{Evaluation Metrics}}: 
We report the solve rate for GSM8K, MATH$_{{lv5*}}$, HumanEval, and MBPP. 
For HotpotQA and DROP, we follow~\cite{zhang2024aflow} and report the F1 Score.
For the industrial case study, we manually compare the accuracy of generated and golden answers.

\subsection{Main Result}
Table~\ref{MainExpTbl} presents the performance of the proposed method across all benchmarks.
Compared to vanilla model, the proposed method achieves an average improvement of more than 8.3\% across the six benchmarks. 
Relative to prior manually designed approaches and automatic workflow optimization methods, our method yields average gains of 14.0\% and 14.6\%, respectively. 
Specifically, compared to AFlow, our self-optimizing hierarchical workflow generation methodology improves performance on MATH$_{{lv5*}}$ benchmarks by an average of 21.2\%. 
Importantly, we achieve these results without relying on validation or test set tuning, demonstrating the effectiveness and self-optimizing capability of our approach across diverse tasks.

\begin{table}[!t]
\tabcolsep = 1.5pt
\centering
\caption{Comparison of performance between vanilla models, manually designed methods and 
automated workflow generation methods for QA, coding, and Math scenarios. We reference and show the performance scores of CoT, CoT SC, MultiPersona, Self Refine, ADAS, and AFlow from~\cite{zhang2024aflow} which are the average scores of three runs.
For vanilla gpt-4o and o1-model runs, we follow ~\cite{zhang2024aflow} and report the average scores of three runs. We run every benchmark once with the proposed method.} 
\vspace{-0.3cm}
\label{MainExpTbl}
\scriptsize{
\begin{tabular}{|c|rr|rr|rr|r|}
\hline
\multirow{2}{*}{Method}                                                                       & \multicolumn{2}{c|}{Multi-Turn QA}                        & \multicolumn{2}{c|}{Coding}                                & \multicolumn{2}{c|}{Math}                              & \multicolumn{1}{c|}{\multirow{2}{*}{Avg.}} \\ \cline{2-7}
                                                                                              & \multicolumn{1}{c|}{HotpotQA} & \multicolumn{1}{c|}{DROP} & \multicolumn{1}{c|}{HumanEval} & \multicolumn{1}{c|}{MBPP} & \multicolumn{1}{c|}{GSM8K} & \multicolumn{1}{c|}{MATH} & \multicolumn{1}{c|}{}                      \\ \hline
\begin{tabular}[c]{@{}c@{}}gpt-4o\\ (Vanilla)\end{tabular}                                    & \multicolumn{1}{r|}{75.0}     & 64.7                      & \multicolumn{1}{r|}{91.5}      & 74.9                      & \multicolumn{1}{r|}{85.5}  & 48.2                      & 73.3                                       \\ \hline
\begin{tabular}[c]{@{}c@{}}o1 model\\ (Vanilla)\end{tabular}                                  & \multicolumn{1}{r|}{70.6}     & 84.9                      & \multicolumn{1}{r|}{89.0}      & 74.5                      & \multicolumn{1}{r|}{94.6}  & 67.1                      & 80.1                                       \\ \hline
\begin{tabular}[c]{@{}c@{}}CoT \\ \cite{wei2022chain}\end{tabular}               & \multicolumn{1}{r|}{67.9}     & 78.5                      & \multicolumn{1}{r|}{88.6}      & 71.8                      & \multicolumn{1}{r|}{92.4}  & 48.8                      & 74.7                                       \\ \hline
\begin{tabular}[c]{@{}c@{}}CoT SC (5-shot)\\ \cite{wang2022self}\end{tabular}    & \multicolumn{1}{r|}{68.9}     & 78.8                      & \multicolumn{1}{r|}{91.6}      & 73.6                      & \multicolumn{1}{r|}{92.7}  & 50.4                      & 76.0                                       \\ \hline
\begin{tabular}[c]{@{}c@{}}MultiPersona\\ \cite{wang2023unleashing}\end{tabular} & \multicolumn{1}{r|}{69.2}     & 74.4                      & \multicolumn{1}{r|}{89.3}      & 73.6                      & \multicolumn{1}{r|}{92.8}  & 50.8                      & 75.1                                       \\ \hline
\begin{tabular}[c]{@{}c@{}}Self Refine\\ \cite{madaan2023self}\end{tabular}      & \multicolumn{1}{r|}{60.8}     & 70.2                      & \multicolumn{1}{r|}{87.8}      & 69.8                      & \multicolumn{1}{r|}{89.6}  & 46.1                      & 70.7                                       \\ \hline
\begin{tabular}[c]{@{}c@{}}ADAS\\ \cite{hu2024automated}\end{tabular}            & \multicolumn{1}{r|}{64.5}     & 76.6                      & \multicolumn{1}{r|}{82.4}      & 53.4                      & \multicolumn{1}{r|}{90.8}  & 35.4                      & 67.2                                       \\ \hline
\begin{tabular}[c]{@{}c@{}}AFlow\\ \cite{zhang2024aflow}\end{tabular}            & \multicolumn{1}{r|}{73.5}     & 80.6                      & \multicolumn{1}{r|}{94.7}      & 83.4                      & \multicolumn{1}{r|}{93.5}  & 56.2                      & 80.3                                       \\ \hline
Proposed                                                                                      & \multicolumn{1}{r|}{81.3}     & 91.8                      & \multicolumn{1}{r|}{99.4}      & 83.1                      & \multicolumn{1}{r|}{97.6}  & 77.4                      & 88.4                                       \\ \hline
\end{tabular}
}
\end{table}

\subsection{Graph, Text, and Code Representation Study}
\begin{figure} [!t]
	\centering
    \includegraphics[width=\columnwidth]{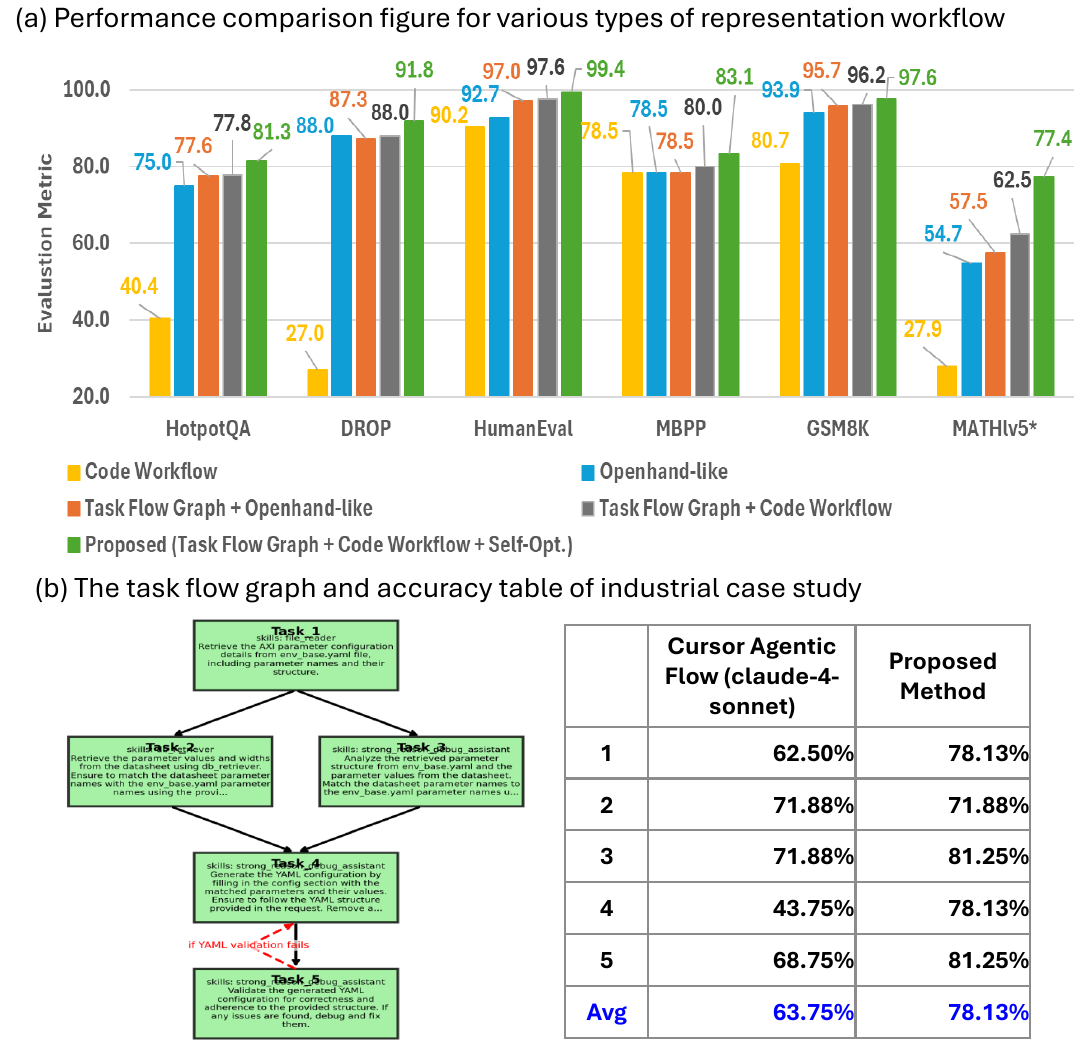}
    \vspace{-0.4cm}
    \caption{(a) Performance comparison of graph, text, and code representation agent study. (b) The example task flow graph topology and accuracy table of industrial case study. Due to confidentiality constraints, we omit details here.}
	\label{AblationIndustryCaseFig}
    \vspace{-0.4cm}
\end{figure}

We implemented a code-represented workflow agent, an Openhand like~\cite{wang2024openhands} text-represented agent, a task flow graph on top of Openhand like agent, and a task flow graph with code-represented workflow agent as described in Section~\ref{PrelimSection} to study the effectiveness of graph, text, and code represented workflows as shown in Figure~\ref{AblationIndustryCaseFig}(a).
The code-represented workflow without optimization perform poorly, primarily because it can not correct errors in the middle of workflow execution.
Adding the task flow graph on top of code-represented workflow significantly improves the evaluation metrics by an average of 30.7\%, as the top-level refining actions (e.g., reruns, jump logic) enabled correction of errors within the flow. 
Additionally, incorporating a task flow graph with the expressiveness of code-represented workflows yielded a further 1.4\% average improvement compared to the OpenHands-like text-represented agent. 
Finally, with the added self-optimizing capability, our proposed method achieved an additional average performance gain of 4.7\%, demonstrating the benefit of integrating hierarchical and self-improving mechanisms.

\subsection{Industrial Case Study}
We extend our evaluation beyond artificial benchmarks to an industrial case in the hardware design domain, which requires digesting multiple files, a design block diagram, and an approximately 100-page datasheet to determine the correct parameter settings for modules with AXI slave/master interfaces~\cite{arm_axi_spec}. 
Previous automatic workflow methods~\cite{hu2024automated, zhang2024aflow} cannot directly applied to this problem since their unknown capability on processing datasheet, and digest multiple files for the problem without valid validation and test datasets.
We ran the proposed method five times, achieving an average accuracy score that is 14.4\% higher than that of the Cursor agentic flow~\cite{anysphere2024cursor}, as shown in Figure~\ref{AblationIndustryCaseFig}(b).

\section{Conclusion}
We propose a self-optimizing agent featuring dynamic hierarchical workflow generation that combines the flexibility of task flow graphs with the expressiveness of subtask-level code-represented workflows. 
Our novel multi-grid-inspired task flow graph optimization, together with a self-reflection-guided EA, dynamically enables effective problem solving without labeled datasets.
The method outperforms existing workflow optimization techniques by an average of 14.6\% across benchmarks in math reasoning, coding, and multi-turn question answering and achieves a 14.4\% higher accuracy than Cursor~\cite{anysphere2024cursor} on a challenging real-world industrial case, demonstrating its effectiveness, self-optimizing capability, and flexibility.

\newpage
\bibliography{aaai2026}

\begin{thebibliography}{66}
\providecommand{\natexlab}[1]{#1}

\bibitem[{{Aider Developers}(2025)}]{aider}
{Aider Developers}. 2025.
\newblock Aider: AI Pair Programming in Your Terminal.
\newblock Accessed: 2025-07-29.

\bibitem[{Anthropic(2024)}]{anthropic2024claude35}
Anthropic. 2024.
\newblock Introducing Claude 3.5 Sonnet.
\newblock \url{https://www.anthropic.com/news/claude-3-5-sonnet}.
\newblock Blog post.

\bibitem[{Anysphere(2024)}]{anysphere2024cursor}
Anysphere. 2024.
\newblock Cursor: The AI Code Editor.
\newblock \url{https://cursor.sh}.
\newblock Version accessed: 2024.

\bibitem[{{ARM Ltd.}(2022)}]{arm_axi_spec}
{ARM Ltd.} 2022.
\newblock \emph{AMBA AXI and ACE Protocol Specification}.
\newblock ARM.
\newblock \url{https://developer.arm.com/documentation/ihi0022/latest}.

\bibitem[{Austin et~al.(2021)Austin, Odena, Nye, Bosma, Michalewski, Dohan, Jiang, Cai, Terry, Le et~al.}]{austin2021program}
Austin, J.; Odena, A.; Nye, M.; Bosma, M.; Michalewski, H.; Dohan, D.; Jiang, E.; Cai, C.; Terry, M.; Le, Q.; et~al. 2021.
\newblock Program synthesis with large language models.
\newblock \emph{arXiv preprint arXiv:2108.07732}.

\bibitem[{Besta et~al.(2024)Besta, Blach, Kubicek, Gerstenberger, Podstawski, Gianinazzi, Gajda, Lehmann, Niewiadomski, Nyczyk et~al.}]{besta2024graph}
Besta, M.; Blach, N.; Kubicek, A.; Gerstenberger, R.; Podstawski, M.; Gianinazzi, L.; Gajda, J.; Lehmann, T.; Niewiadomski, H.; Nyczyk, P.; et~al. 2024.
\newblock Graph of thoughts: Solving elaborate problems with large language models.
\newblock In \emph{Proceedings of the AAAI conference on artificial intelligence}, volume~38, 17682--17690.

\bibitem[{Chang et~al.(2025)Chang, Ho, Li, Chen, and Ren}]{chang2025drc}
Chang, C.-C.; Ho, C.-T.; Li, Y.; Chen, Y.; and Ren, H. 2025.
\newblock DRC-Coder: Automated drc checker code generation using LLM autonomous agent.
\newblock In \emph{Proceedings of the 2025 International Symposium on Physical Design}, 143--151.

\bibitem[{Chen et~al.(2021)Chen, Tworek, Jun, Yuan, Pinto, Kaplan, Edwards, Burda, Joseph, Brockman et~al.}]{chen2021evaluating}
Chen, M.; Tworek, J.; Jun, H.; Yuan, Q.; Pinto, H. P. D.~O.; Kaplan, J.; Edwards, H.; Burda, Y.; Joseph, N.; Brockman, G.; et~al. 2021.
\newblock Evaluating large language models trained on code.
\newblock \emph{arXiv preprint arXiv:2107.03374}.

\bibitem[{Cobbe et~al.(2021)Cobbe, Kosaraju, Bavarian, Chen, Jun, Kaiser, Plappert, Tworek, Hilton, Nakano et~al.}]{cobbe2021training}
Cobbe, K.; Kosaraju, V.; Bavarian, M.; Chen, M.; Jun, H.; Kaiser, L.; Plappert, M.; Tworek, J.; Hilton, J.; Nakano, R.; et~al. 2021.
\newblock Training verifiers to solve math word problems.
\newblock \emph{arXiv preprint arXiv:2110.14168}.

\bibitem[{Dua et~al.(2019)Dua, Wang, Dasigi, Stanovsky, Singh, and Gardner}]{dua2019drop}
Dua, D.; Wang, Y.; Dasigi, P.; Stanovsky, G.; Singh, S.; and Gardner, M. 2019.
\newblock DROP: A reading comprehension benchmark requiring discrete reasoning over paragraphs.
\newblock \emph{arXiv preprint arXiv:1903.00161}.

\bibitem[{Fernando et~al.(2023)Fernando, Banarse, Michalewski, Osindero, and Rockt{\"a}schel}]{fernando2023promptbreeder}
Fernando, C.; Banarse, D.; Michalewski, H.; Osindero, S.; and Rockt{\"a}schel, T. 2023.
\newblock Promptbreeder: Self-referential self-improvement via prompt evolution.
\newblock \emph{arXiv preprint arXiv:2309.16797}.

\bibitem[{Hendrycks et~al.(2021)Hendrycks, Burns, Kadavath, Arora, Basart, Tang, Song, and Steinhardt}]{hendrycks2021measuring}
Hendrycks, D.; Burns, C.; Kadavath, S.; Arora, A.; Basart, S.; Tang, E.; Song, D.; and Steinhardt, J. 2021.
\newblock Measuring mathematical problem solving with the math dataset.
\newblock \emph{arXiv preprint arXiv:2103.03874}.

\bibitem[{Ho et~al.(2025)Ho, Ren, Khailany et~al.}]{ho2025verilogcoder}
Ho, C.-T.; Ren, H.; Khailany, B.; et~al. 2025.
\newblock Verilogcoder: Autonomous verilog coding agents with graph-based planning and abstract syntax tree (ast)-based waveform tracing tool.
\newblock In \emph{Proceedings of the AAAI Conference on Artificial Intelligence}, volume~39, 300--307.

\bibitem[{Hong et~al.(2024)Hong, Lin, Liu, Liu, Wu, Zhang, Wei, Li, Chen, Zhang et~al.}]{hong2024data}
Hong, S.; Lin, Y.; Liu, B.; Liu, B.; Wu, B.; Zhang, C.; Wei, C.; Li, D.; Chen, J.; Zhang, J.; et~al. 2024.
\newblock Data interpreter: An llm agent for data science.
\newblock \emph{arXiv preprint arXiv:2402.18679}.

\bibitem[{Hu et~al.(2023)}]{hu2023thought}
Hu, C.~J., Shengran; et~al. 2023.
\newblock Thought cloning: Learning to think while acting by imitating human thinking.
\newblock \emph{Advances in Neural Information Processing Systems}, 36: 44451--44469.

\bibitem[{Hu et~al.(2024)}]{hu2024automated}
Hu, L. C. C.~J., Shengran; et~al. 2024.
\newblock Automated design of agentic systems.
\newblock \emph{arXiv preprint arXiv:2408.08435}.

\bibitem[{Huang et~al.(2023)Huang, Bu, Zhang, Luck, and Cui}]{huang2023agentcoder}
Huang, D.; Bu, Q.; Zhang, J.~M.; Luck, M.; and Cui, H. 2023.
\newblock Agentcoder: Multi-agent-based code generation with iterative testing and optimisation.
\newblock \emph{arXiv preprint arXiv:2312.13010}.

\bibitem[{Jaech et~al.(2024)Jaech, Kalai, Lerer, Richardson, El-Kishky, Low, Helyar, Madry, Beutel, Carney et~al.}]{jaech2024openai}
Jaech, A.; Kalai, A.; Lerer, A.; Richardson, A.; El-Kishky, A.; Low, A.; Helyar, A.; Madry, A.; Beutel, A.; Carney, A.; et~al. 2024.
\newblock Openai o1 system card.
\newblock \emph{arXiv preprint arXiv:2412.16720}.

\bibitem[{Karypis, Kumar et~al.(1999)}]{karypis1999multilevel}
Karypis, G.; Kumar, V.; et~al. 1999.
\newblock Multilevel k-way hypergraph partitioning.
\newblock In \emph{Proceedings of the 36th annual ACM/IEEE design automation conference}, 343--348.

\bibitem[{Khattab et~al.(2024)Khattab, Singhvi, Maheshwari, Zhang, Santhanam, Haq, Sharma, Joshi, Moazam, Miller et~al.}]{khattab2024dspy}
Khattab, O.; Singhvi, A.; Maheshwari, P.; Zhang, Z.; Santhanam, K.; Haq, S.; Sharma, A.; Joshi, T.~T.; Moazam, H.; Miller, H.; et~al. 2024.
\newblock Dspy: Compiling declarative language model calls into state-of-the-art pipelines.
\newblock In \emph{The Twelfth International Conference on Learning Representations}.

\bibitem[{Lai et~al.(2025)Lai, Lee, Chen, Poddar, Hu, Pan, and Luo}]{lai2025analogcoder}
Lai, Y.; Lee, S.; Chen, G.; Poddar, S.; Hu, M.; Pan, D.~Z.; and Luo, P. 2025.
\newblock Analogcoder: Analog circuit design via training-free code generation.
\newblock In \emph{Proceedings of the AAAI Conference on Artificial Intelligence}, volume~39, 379--387.

\bibitem[{Le, Zhang et~al.(2023)}]{le2023log}
Le, V.-H.; Zhang, H.; et~al. 2023.
\newblock Log parsing with prompt-based few-shot learning.
\newblock In \emph{2023 IEEE/ACM 45th International Conference on Software Engineering (ICSE)}, 2438--2449. IEEE.

\bibitem[{Lewis et~al.(2020)Lewis, Perez, Piktus, Petroni, Karpukhin, Goyal, K{\"u}ttler, Lewis, Yih, Rockt{\"a}schel et~al.}]{lewis2020retrieval}
Lewis, P.; Perez, E.; Piktus, A.; Petroni, F.; Karpukhin, V.; Goyal, N.; K{\"u}ttler, H.; Lewis, M.; Yih, W.-t.; Rockt{\"a}schel, T.; et~al. 2020.
\newblock Retrieval-augmented generation for knowledge-intensive nlp tasks.
\newblock \emph{Advances in neural information processing systems}, 33: 9459--9474.

\bibitem[{Li et~al.(2024)Li, Xu, Mei, Hua, Rama, Raheja, Wang, Zhu, and Zhang}]{li2024autoflow}
Li, Z.; Xu, S.; Mei, K.; Hua, W.; Rama, B.; Raheja, O.; Wang, H.; Zhu, H.; and Zhang, Y. 2024.
\newblock Autoflow: Automated workflow generation for large language model agents.
\newblock \emph{arXiv preprint arXiv:2407.12821}.

\bibitem[{Liu et~al.(2025)Liu, Fang, Zhou, Wang, and Meng}]{liu2025sew}
Liu, S.; Fang, J.; Zhou, H.; Wang, Y.; and Meng, Z. 2025.
\newblock SEW: Self-Evolving Agentic Workflows for Automated Code Generation.
\newblock \emph{arXiv preprint arXiv:2505.18646}.

\bibitem[{Liu et~al.(2022)Liu, Zhang, He, Zhang, Li, Kang, Xu, Ma, Lin, Dang et~al.}]{liu2022uniparser}
Liu, Y.; Zhang, X.; He, S.; Zhang, H.; Li, L.; Kang, Y.; Xu, Y.; Ma, M.; Lin, Q.; Dang, Y.; et~al. 2022.
\newblock Uniparser: A unified log parser for heterogeneous log data.
\newblock In \emph{Proceedings of the ACM Web Conference 2022}, 1893--1901.

\bibitem[{Liu et~al.(2023)Liu, Zhang, Li, Liu, and Yang}]{liu2023dynamic}
Liu, Z.; Zhang, Y.; Li, P.; Liu, Y.; and Yang, D. 2023.
\newblock Dynamic llm-agent network: An llm-agent collaboration framework with agent team optimization.
\newblock \emph{arXiv preprint arXiv:2310.02170}.

\bibitem[{Madaan et~al.(2023)Madaan, Tandon, Gupta, Hallinan, Gao, Wiegreffe, Alon, Dziri, Prabhumoye, Yang et~al.}]{madaan2023self}
Madaan, A.; Tandon, N.; Gupta, P.; Hallinan, S.; Gao, L.; Wiegreffe, S.; Alon, U.; Dziri, N.; Prabhumoye, S.; Yang, Y.; et~al. 2023.
\newblock Self-refine: Iterative refinement with self-feedback.
\newblock \emph{Advances in Neural Information Processing Systems}, 36: 46534--46594.

\bibitem[{Mouret, Clune et~al.(2015)}]{mouret2015illuminating}
Mouret, J.-B.; Clune, J.; et~al. 2015.
\newblock Illuminating search spaces by mapping elites.
\newblock \emph{arXiv preprint arXiv:1504.04909}.

\bibitem[{Nainani et~al.(2025)Nainani, Ho, Dhurka, and Ren}]{nainani2025timing}
Nainani, J.; Ho, C.-T.; Dhurka, A.; and Ren, H. 2025.
\newblock Timing Analysis Agent: Autonomous Multi-Corner Multi-Mode (MCMM) Timing Debugging with Timing Debug Relation Graph.
\newblock \emph{arXiv preprint arXiv:2504.11502}.

\bibitem[{Novikov et~al.(2025)Novikov, V{\~u}, Eisenberger, Dupont, Huang, Wagner, Shirobokov, Kozlovskii, Ruiz, Mehrabian et~al.}]{novikov2025alphaevolve}
Novikov, A.; V{\~u}, N.; Eisenberger, M.; Dupont, E.; Huang, P.-S.; Wagner, A.~Z.; Shirobokov, S.; Kozlovskii, B.; Ruiz, F.~J.; Mehrabian, A.; et~al. 2025.
\newblock AlphaEvolve: A coding agent for scientific and algorithmic discovery.
\newblock \emph{arXiv preprint arXiv:2506.13131}.

\bibitem[{{OpenAI}({2023})}]{GPT4}
{OpenAI}. {2023}.
\newblock {Gpt-4 technical report}.

\bibitem[{OpenAI(2024{\natexlab{a}})}]{openai2024gpt4o}
OpenAI. 2024{\natexlab{a}}.
\newblock GPT-4o-1120.
\newblock \url{https://platform.openai.com/docs/models/gpt-4o}.
\newblock \url{https://openai.com/research/gpt-4o}.

\bibitem[{OpenAI(2024{\natexlab{b}})}]{openai2024o1}
OpenAI. 2024{\natexlab{b}}.
\newblock OpenAI o1 model.
\newblock Available at \url{https://platform.openai.com/docs/models}.

\bibitem[{Romera-Paredes et~al.(2024)Romera-Paredes, Barekatain, Novikov, Balog, Kumar, Dupont, Ruiz, Ellenberg, Wang, Fawzi et~al.}]{romera2024mathematical}
Romera-Paredes, B.; Barekatain, M.; Novikov, A.; Balog, M.; Kumar, M.~P.; Dupont, E.; Ruiz, F.~J.; Ellenberg, J.~S.; Wang, P.; Fawzi, O.; et~al. 2024.
\newblock Mathematical discoveries from program search with large language models.
\newblock \emph{Nature}, 625(7995): 468--475.

\bibitem[{Saad-Falcon et~al.(2024)Saad-Falcon, Lafuente, Natarajan, Maru, Todorov, Guha, Buchanan, Chen, Guha, R{\'e} et~al.}]{saad2024archon}
Saad-Falcon, J.; Lafuente, A.~G.; Natarajan, S.; Maru, N.; Todorov, H.; Guha, E.; Buchanan, E.~K.; Chen, M.; Guha, N.; R{\'e}, C.; et~al. 2024.
\newblock Archon: An architecture search framework for inference-time techniques.
\newblock \emph{arXiv preprint arXiv:2409.15254}.

\bibitem[{Schick et~al.(2023)Schick, Dwivedi-Yu, Dess{\`\i}, Raileanu, Lomeli, Hambro, Zettlemoyer, Cancedda, and Scialom}]{schick2023toolformer}
Schick, T.; Dwivedi-Yu, J.; Dess{\`\i}, R.; Raileanu, R.; Lomeli, M.; Hambro, E.; Zettlemoyer, L.; Cancedda, N.; and Scialom, T. 2023.
\newblock Toolformer: Language models can teach themselves to use tools.
\newblock \emph{Advances in Neural Information Processing Systems}, 36: 68539--68551.

\bibitem[{Sharma(2025)}]{openevolve}
Sharma, A. 2025.
\newblock OpenEvolve: an open-source evolutionary coding agent.

\bibitem[{Shinn et~al.(2023)Shinn, Cassano, Gopinath, Narasimhan, and Yao}]{shinn2023reflexion}
Shinn, N.; Cassano, F.; Gopinath, A.; Narasimhan, K.; and Yao, S. 2023.
\newblock Reflexion: Language agents with verbal reinforcement learning.
\newblock \emph{Advances in Neural Information Processing Systems}, 36: 8634--8652.

\bibitem[{Sohrabizadeh et~al.()Sohrabizadeh, Song, Liu, Roy, Lee, Raiman, and Catanzaro}]{sohrabizadehnemotron}
Sohrabizadeh, A.; Song, J.; Liu, M.; Roy, R.; Lee, C.; Raiman, J.; and Catanzaro, B. ????
\newblock Nemotron-CORTEXA: Enhancing LLM Agents for Software Engineering Tasks via Improved Localization and Solution Diversity.
\newblock In \emph{Forty-second International Conference on Machine Learning}.

\bibitem[{Sun et~al.(2023)Sun, Xu, Liu, Luan, Wang, Shang, Wen, and Yan}]{sun2023indeterminacy}
Sun, H.; Xu, W.; Liu, W.; Luan, J.; Wang, B.; Shang, S.; Wen, J.-R.; and Yan, R. 2023.
\newblock From indeterminacy to determinacy: Augmenting logical reasoning capabilities with large language models.

\bibitem[{Tanese(1989)}]{tanese1989distributed}
Tanese, R. 1989.
\newblock \emph{Distributed genetic algorithms for function optimization}.
\newblock University of Michigan.

\bibitem[{Tang et~al.(2023)Tang, Yang, Fan, Cao, Luo, and Halevy}]{tang2023verifai}
Tang, N.; Yang, C.; Fan, J.; Cao, L.; Luo, Y.; and Halevy, A. 2023.
\newblock VerifAI: verified generative AI.
\newblock \emph{arXiv preprint arXiv:2307.02796}.

\bibitem[{Trottenberg et~al.(2001)Trottenberg, Oosterlee, Schuller et~al.}]{trottenberg2001multigrid}
Trottenberg, U.; Oosterlee, C.~W.; Schuller, A.; et~al. 2001.
\newblock \emph{Multigrid methods}.
\newblock Academic press.

\bibitem[{Wang et~al.(2024{\natexlab{a}})Wang, Ma, Feng, Zhang, Yang, Zhang, Chen, Tang, Chen, Lin et~al.}]{wang2024survey}
Wang, L.; Ma, C.; Feng, X.; Zhang, Z.; Yang, H.; Zhang, J.; Chen, Z.; Tang, J.; Chen, X.; Lin, Y.; et~al. 2024{\natexlab{a}}.
\newblock A survey on large language model based autonomous agents.
\newblock \emph{Frontiers of Computer Science}, 18(6): 186345.

\bibitem[{Wang et~al.(2024{\natexlab{b}})Wang, Li, Song, Xu, Tang, Zhuge, Pan, Song, Li, Singh et~al.}]{wang2024openhands}
Wang, X.; Li, B.; Song, Y.; Xu, F.~F.; Tang, X.; Zhuge, M.; Pan, J.; Song, Y.; Li, B.; Singh, J.; et~al. 2024{\natexlab{b}}.
\newblock Openhands: An open platform for ai software developers as generalist agents.
\newblock \emph{arXiv preprint arXiv:2407.16741}.

\bibitem[{Wang et~al.(2022)Wang, Wei, Schuurmans, Le, Chi, Narang, Chowdhery, and Zhou}]{wang2022self}
Wang, X.; Wei, J.; Schuurmans, D.; Le, Q.; Chi, E.; Narang, S.; Chowdhery, A.; and Zhou, D. 2022.
\newblock Self-consistency improves chain of thought reasoning in language models.
\newblock \emph{arXiv preprint arXiv:2203.11171}.

\bibitem[{Wang et~al.(2023)Wang, Mao, Wu, Ge, Wei, and Ji}]{wang2023unleashing}
Wang, Z.; Mao, S.; Wu, W.; Ge, T.; Wei, F.; and Ji, H. 2023.
\newblock Unleashing the emergent cognitive synergy in large language models: A task-solving agent through multi-persona self-collaboration.
\newblock \emph{arXiv preprint arXiv:2307.05300}.

\bibitem[{Wei et~al.(2022)Wei, Wang, Schuurmans, Bosma, Xia, Chi, Le, Zhou et~al.}]{wei2022chain}
Wei, J.; Wang, X.; Schuurmans, D.; Bosma, M.; Xia, F.; Chi, E.; Le, Q.~V.; Zhou, D.; et~al. 2022.
\newblock Chain-of-thought prompting elicits reasoning in large language models.
\newblock \emph{Advances in neural information processing systems}, 35: 24824--24837.

\bibitem[{Weng(2023)}]{weng2023agent}
Weng, L. 2023.
\newblock LLM-powered Autonomous Agents.
\newblock \emph{lilianweng.github.io}.

\bibitem[{Wu et~al.(2023)Wu, Bansal, Zhang, Wu, Zhang, Zhu, Li, Jiang, Zhang, and Wang}]{wu2023autogen}
Wu, Q.; Bansal, G.; Zhang, J.; Wu, Y.; Zhang, S.; Zhu, E.; Li, B.; Jiang, L.; Zhang, X.; and Wang, C. 2023.
\newblock Autogen: Enabling next-gen llm applications via multi-agent conversation framework.
\newblock \emph{arXiv preprint arXiv:2308.08155}.

\bibitem[{Xia et~al.(2024)Xia, Deng, Dunn, and Zhang}]{xia2024agentless}
Xia, C.~S.; Deng, Y.; Dunn, S.; and Zhang, L. 2024.
\newblock Agentless: Demystifying llm-based software engineering agents.
\newblock \emph{arXiv preprint arXiv:2407.01489}.

\bibitem[{Yang et~al.(2023)Yang, Wang, Lu, Liu, Le, Zhou, and Chen}]{yang2023large}
Yang, C.; Wang, X.; Lu, Y.; Liu, H.; Le, Q.~V.; Zhou, D.; and Chen, X. 2023.
\newblock Large language models as optimizers.
\newblock In \emph{The Twelfth International Conference on Learning Representations}.

\bibitem[{Yang et~al.(2024{\natexlab{a}})Yang, Jimenez, Wettig, Lieret, Yao, Narasimhan, and Press}]{yang2024swe}
Yang, J.; Jimenez, C.~E.; Wettig, A.; Lieret, K.; Yao, S.; Narasimhan, K.; and Press, O. 2024{\natexlab{a}}.
\newblock Swe-agent: Agent-computer interfaces enable automated software engineering.
\newblock \emph{arXiv preprint arXiv:2405.15793}.

\bibitem[{Yang et~al.(2024{\natexlab{b}})Yang, Yu, Zhang, Cao, Xu, Zhang, Gonzalez, and Cui}]{yang2024buffer}
Yang, L.; Yu, Z.; Zhang, T.; Cao, S.; Xu, M.; Zhang, W.; Gonzalez, J.~E.; and Cui, B. 2024{\natexlab{b}}.
\newblock Buffer of thoughts: Thought-augmented reasoning with large language models.
\newblock \emph{Advances in Neural Information Processing Systems}, 37: 113519--113544.

\bibitem[{Yang et~al.(2018)Yang, Qi, Zhang, Bengio, Cohen, Salakhutdinov, and Manning}]{yang2018hotpotqa}
Yang, Z.; Qi, P.; Zhang, S.; Bengio, Y.; Cohen, W.~W.; Salakhutdinov, R.; and Manning, C.~D. 2018.
\newblock HotpotQA: A dataset for diverse, explainable multi-hop question answering.
\newblock \emph{arXiv preprint arXiv:1809.09600}.

\bibitem[{Yao et~al.(2023)Yao, Yu, Zhao, Shafran, Griffiths, Cao, and Narasimhan}]{yao2023tree}
Yao, S.; Yu, D.; Zhao, J.; Shafran, I.; Griffiths, T.; Cao, Y.; and Narasimhan, K. 2023.
\newblock Tree of thoughts: Deliberate problem solving with large language models.
\newblock \emph{Advances in neural information processing systems}, 36: 11809--11822.

\bibitem[{Yao et~al.(2022)Yao, Zhao, Yu, Du, Shafran, Narasimhan, and Cao}]{yao2022react}
Yao, S.; Zhao, J.; Yu, D.; Du, N.; Shafran, I.; Narasimhan, K.; and Cao, Y. 2022.
\newblock React: Synergizing reasoning and acting in language models.
\newblock \emph{arXiv preprint arXiv:2210.03629}.

\bibitem[{Yu, He, and Ying(2023)}]{yu2023thought}
Yu, J.; He, R.; and Ying, R. 2023.
\newblock Thought propagation: An analogical approach to complex reasoning with large language models.
\newblock \emph{arXiv preprint arXiv:2310.03965}.

\bibitem[{Yuksekgonul et~al.(2024)Yuksekgonul, Bianchi, Boen, Liu, Huang, Guestrin, and Zou}]{yuksekgonul2024textgrad}
Yuksekgonul, M.; Bianchi, F.; Boen, J.; Liu, S.; Huang, Z.; Guestrin, C.; and Zou, J. 2024.
\newblock Textgrad: Automatic" differentiation" via text.
\newblock \emph{arXiv preprint arXiv:2406.07496}.

\bibitem[{Zhang et~al.(2025)Zhang, Chen, Wan, Chang, Cheng, Wang, Hu, and Bai}]{zhang2025evoflow}
Zhang, G.; Chen, K.; Wan, G.; Chang, H.; Cheng, H.; Wang, K.; Hu, S.; and Bai, L. 2025.
\newblock Evoflow: Evolving diverse agentic workflows on the fly.
\newblock \emph{arXiv preprint arXiv:2502.07373}.

\bibitem[{Zhang et~al.(2024{\natexlab{a}})Zhang, Xiang, Yu, Teng, Chen, Chen, Zhuge, Cheng, Hong, Wang et~al.}]{zhang2024aflow}
Zhang, J.; Xiang, J.; Yu, Z.; Teng, F.; Chen, X.; Chen, J.; Zhuge, M.; Cheng, X.; Hong, S.; Wang, J.; et~al. 2024{\natexlab{a}}.
\newblock Aflow: Automating agentic workflow generation.
\newblock \emph{arXiv preprint arXiv:2410.10762}.

\bibitem[{Zhang et~al.(2024{\natexlab{b}})Zhang, Dai, Bo, Ma, Li, Chen, Zhu, Dong, and Wen}]{zhang2024survey}
Zhang, Z.; Dai, Q.; Bo, X.; Ma, C.; Li, R.; Chen, X.; Zhu, J.; Dong, Z.; and Wen, J.-R. 2024{\natexlab{b}}.
\newblock A survey on the memory mechanism of large language model based agents.
\newblock \emph{ACM Transactions on Information Systems}.

\bibitem[{Zhong et~al.(2024)Zhong, Wang, Xu, Liu, Ding, and Du}]{zhong2024achieving}
Zhong, Q.; Wang, K.; Xu, Z.; Liu, J.; Ding, L.; and Du, B. 2024.
\newblock Achieving> 97\% on gsm8k: Deeply understanding the problems makes llms better solvers for math word problems.
\newblock \emph{arXiv preprint arXiv:2404.14963}.

\bibitem[{Zhou et~al.(2024)Zhou, Ou, Ding, Li, Wu, Wang, Chen, Wang, Xu, Zhang et~al.}]{zhou2024symbolic}
Zhou, W.; Ou, Y.; Ding, S.; Li, L.; Wu, J.; Wang, T.; Chen, J.; Wang, S.; Xu, X.; Zhang, N.; et~al. 2024.
\newblock Symbolic learning enables self-evolving agents.
\newblock \emph{arXiv preprint arXiv:2406.18532}.

\bibitem[{Zhuge et~al.(2024)Zhuge, Wang, Kirsch, Faccio, Khizbullin, and Schmidhuber}]{zhuge2024gptswarm}
Zhuge, M.; Wang, W.; Kirsch, L.; Faccio, F.; Khizbullin, D.; and Schmidhuber, J. 2024.
\newblock Gptswarm: Language agents as optimizable graphs.
\newblock In \emph{Forty-first International Conference on Machine Learning}.

\end{thebibliography}

\newpage

\onecolumn


\section*{Supplementary material}

# Appendix

Anonymous submission

## Input Task Prompt for Multi-Turn QA, MATH, and Coding Benchmarks

We show the input task prompt for six public benchmarks: (1) Multi-turn QA: HotpotQA (Yang et al. 2018) and DROP (Dua et al. 2019); (2) Coding: HumanEval (Chen et al. 2021) and MBPP (Austin et al. 2021); (3) Math: MATH$_{lv5*}$ (Hendrycks et al. 2021; Hong et al. 2024) and GSM8K (Cobbe et al. 2021), in Figure 1. We follow (Zhang et al. 2024) on output format definition and evaluation method for automatic evaluation on the benchmarks.

---

**MATH Reasoning: Input Task Prompt of MATHlv5***

You need to calculate the final answer for the problem.

[Output Format]:
You must reply the final answer (i.e., a number) in the $\boxed\{{Number\}}$ format only. For fraction number, use \\frac{num}{delim} to express. For example, $\boxed{{-\\frac{{1}}{{16}}}}$

[Question]:
{Question}

---

**MATH Reasoning: Input Task Prompt of GSM8K**

You need to reply the number that you calculated.

[Output Format]:
Reply the final answer in the **Number** in string format only.

[Question]:
{Question}

---

**Coding: Input Task Prompt of HumanEval**

You are n Python Coding expert.

[Output Format]:
Only return the completed {FunctionName} function in ```python and ``` code block!

[Question]:
{Question}

---

**Coding: Input Task Prompt of MBPP**

You are n Python Coding expert.

[Output Format]:
Only return the completed {FunctionName} function in ```python and ``` code block!

[Question]:
{Question}

---

**Multi-turn QA: Input Task Prompt of HotpotQA**

Write a high-quality answer for the given question. You should come out the answer in **Answer** without irrelevant sentences and words.

### Below are examples:
Example 1:
Document [1](Title: Stuart Rosenberg) Stuart Rosenberg (August 11, 1927 – March 15, 2007) was an American film and television director whose motion pictures include "Cool Hand Luke" (1967), "Voyage of the Damned" (1976), "The Amityville Horror" (1979), and "The Pope of Greenwich Village" (1984).He was noted for his work with actor Paul Newman.

[Partial example content omitted due to space limit]
Question: Are director of film Move (1970 Film) and director of film Méditerranée (1963 Film) from the same country?
Answer: **no**

Example 2:
Document [1](Title: Pamela Jain) Pamela Jain is an Indian playback singer.Date of Birth:16th March.

[Partial example content omitted due to space limit]
Question: What is the date of birth of Mina Gerhardsen's father?
Answer: **13 June 1946**

Question: {Question}
Answer:

---

**Multi-turn QA: Input Task Prompt of DROP**

Write a high-quality answer for the given. You should come out the answer in **Answer** without irrelevant sentences and words.

[Example 1]:
Passage: The population consisted of 5,841 people (25.3%) under age 18, 1,875 people (8.1%) age 18 to 24,

[Partial example content omitted due to space limit]
Question: How many more people were age 25 to 44 than age 18 to 24?
Answer: **3150**

[Example 2]:
Passage: In the county, the population was spread out with 25.00% under the age of 18, 17.10% from 18 to 24,

[Partial example content omitted due to space limit]
Question: Which age group is larger: 25 to 44 or 45 to 64?
Answer: **25 to 44**

Question: {Question}
Answer:

Figure 1: The input task prompt of Multi-Turn QA, MATH, and Coding Benchmarks.

# Task Flow Graph Optimization Study

We study the statistics of task flow graph with and without applying multi-grid-inspired optimization methodology. Table 1 shows the the statistics of the task flow graphs before and after multi-grid–inspired task flow graph optimization. The effective score for a given graph $G(T, E)$ is calculated as:

$$S_g = \frac{1}{|T|} \sum_{i=1}^{|T|} s_i, \tag{1}$$

where $|T|$ is the number of subtasks in the task flow graph and $s_i$ is the effective score of subtask node $i$. We observe that the average number of nodes and edges in the optimized task flow graphs is reduced, improving efficiency, while the average standard deviation of nodes and edges slightly increases, reflecting the need to handle problems of varying complexity. In addition, the average $S_g$ increases by 4.67% after task flow graph optimization. For $\text{MATH}_{\text{lv5}*}$, $S_g$ improves by up to 8%, which aligns with our proposed method outperforming the baselines by up to 21.2%. Figure 2 shows the statistical distribution of $S_g$. These results demonstrate that the distribution of $S_g$ shifts to the right across the Multi-Turn QA, Coding, and Math reasoning benchmarks, confirming that the proposed graph optimization method effectively improves the objective function.

Table 1: The statistics of task flow graph before and after multi-grid-inspired task flow graph optimization. $Edge_d$ and $Edge_j$ represent the dependency edge and jump logic edge, repectively. $S_g$ represents the effective score of the task flow graph. avg and std are the average and the standard deviation, respectively.

| Benchmark | Original Task Flow Graph | | | | | | | | Optimized Task Flow Graph | | | | | | | | Impr. Avg. $S_g$ |
|---|---|---|---|---|---|---|---|---|---|---|---|---|---|---|---|---|---|
| | $Node$ | | $Edge_d$ | | $Edge_j$ | | $S_g$ | | $Node$ | | $Edge_d$ | | $Edge_j$ | | $S_g$ | | |
| | avg | std | avg | std | avg | std | avg | std | avg | std | avg | std | avg | std | avg | std | |
| HotPotQA | 2.67 | 0.70 | 1.56 | 0.68 | 0.40 | 0.58 | 0.19 | 0.05 | 1.60 | 0.97 | 0.46 | 0.94 | 0.75 | 0.90 | 0.23 | 0.07 | 4% |
| DROP | 2.52 | 0.58 | 1.45 | 0.59 | 0.07 | 0.26 | 0.14 | 0.04 | 1.42 | 0.80 | 0.36 | 0.74 | 0.75 | 0.81 | 0.17 | 0.07 | 3% |
| HumanEval | 3.99 | 0.53 | 3.50 | 0.92 | 1.48 | 0.72 | 0.21 | 0.04 | 1.76 | 1.05 | 0.72 | 1.21 | 1.13 | 0.80 | 0.28 | 0.07 | 7% |
| MBPP | 3.03 | 0.31 | 2.06 | 0.48 | 1.43 | 0.62 | 0.19 | 0.05 | 1.53 | 0.99 | 0.57 | 1.17 | 0.84 | 0.72 | 0.23 | 0.08 | 4% |
| GSM8K | 2.10 | 0.81 | 1.04 | 0.86 | 0.04 | 0.19 | 0.13 | 0.04 | 1.24 | 0.51 | 0.17 | 0.47 | 0.67 | 0.76 | 0.15 | 0.06 | 2% |
| $\text{MATH}_{\text{lv5}*}$ | 2.83 | 0.77 | 1.99 | 0.93 | 0.72 | 0.82 | 0.22 | 0.07 | 1.34 | 0.75 | 0.32 | 0.82 | 0.84 | 0.74 | 0.30 | 0.09 | 8% |
| Avg. | 2.86 | 0.62 | 1.93 | 0.74 | 0.69 | 0.53 | 0.18 | 0.05 | 1.48 | 0.85 | 0.43 | 0.89 | 0.83 | 0.79 | 0.23 | 0.07 | 4.67% |

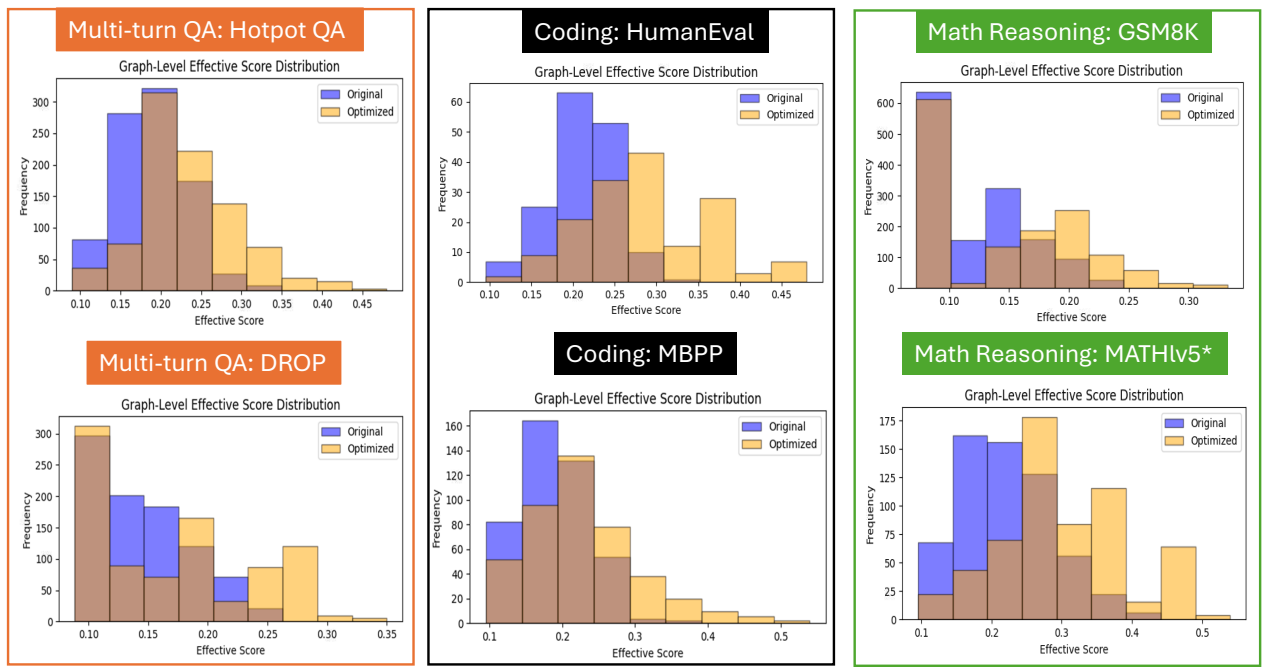

Figure 2: The statistic distribution of task flow graph before and after multi-grid-inspired task flow graph optimization on HotpotQA, DROP, HumanEval, MBPP, GSM8k, and $\text{MATH}_{\text{lv5}*}$.

We present two examples of task flow graph optimization from the HumanEval benchmarks in Figure 3. For HumanEval/15, the optimized task flow graph is simplified and executed more efficiently, as the problem is considered relatively easy. In addition, based on prior knowledge that the coder assistant writes and executes code together, the writing and execution subtasks are merged for improved efficiency. In contrast, HumanEval/69 is considered a more complex task based on its problem description. Therefore, the optimized task flow graph ensures that all reasoning subtasks are completed before writing the Python code and handling corner cases. These examples demonstrate that the proposed method efficiently optimizes task flow graphs according to the complexity and dynamic requirements of the problems. In addition, we demonstrate the optimized task flow examples for MATH reasoning (i.e., MATH$_{lv5*}$) and multi-turn QA (i.e., HotPotQA) in Figrue 4 and Figure 5, respectively. The optimized task flow graph incorporates additional information extraction and reasoning steps before generating the final answer for the multi-turn QA task.

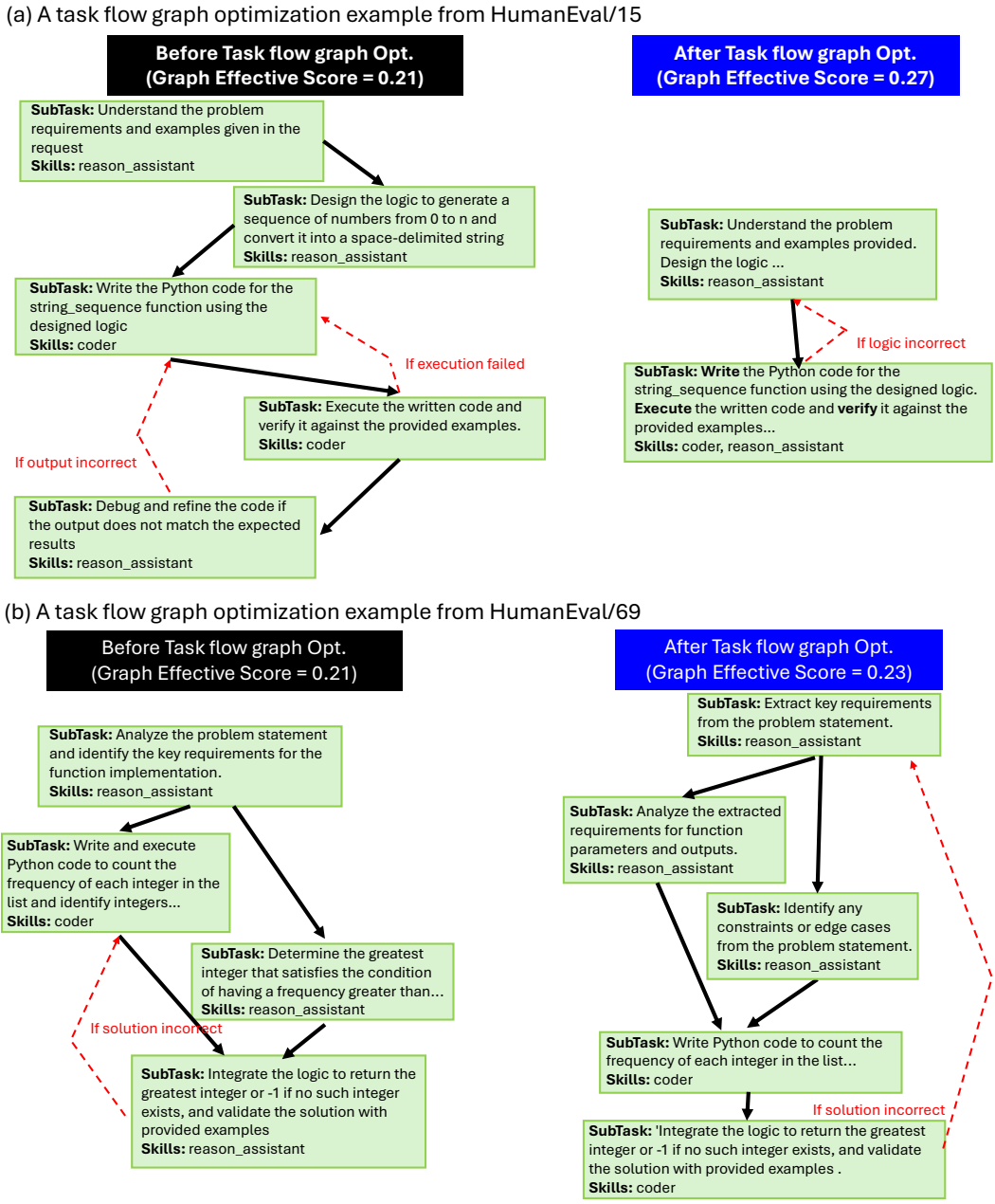

Figure 3: Two task flow graph optimization examples from HumanEval benchmarks.

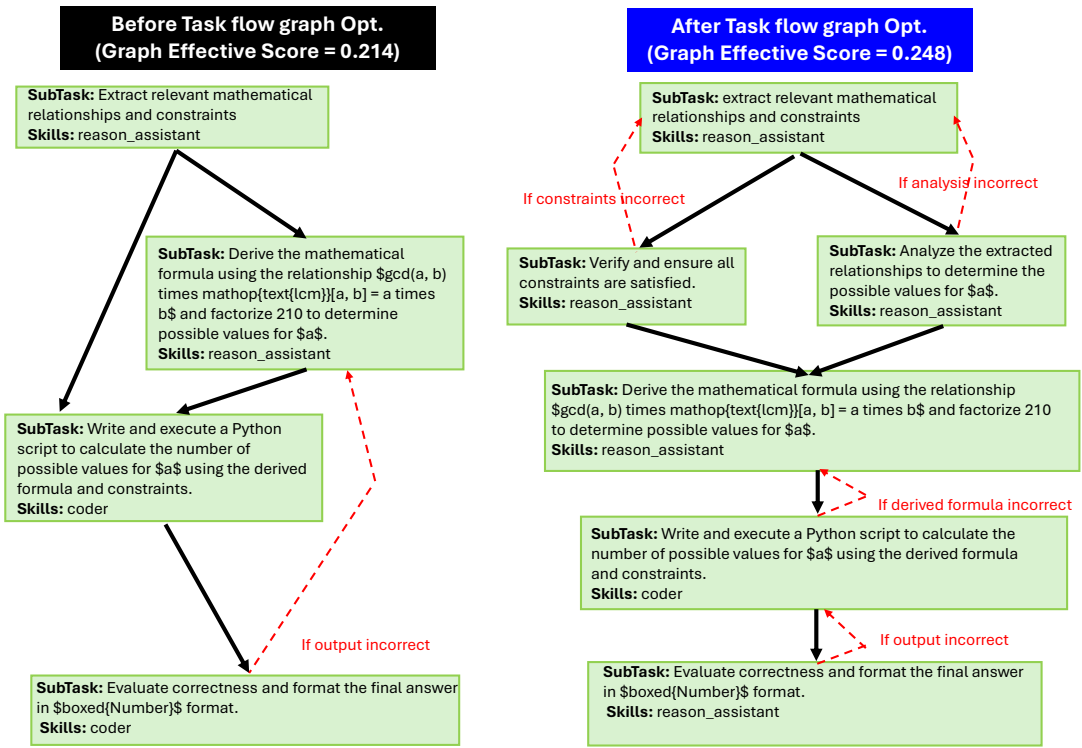

Figure 4: A task flow graph optimization example from MATH$_{lv5*}$.

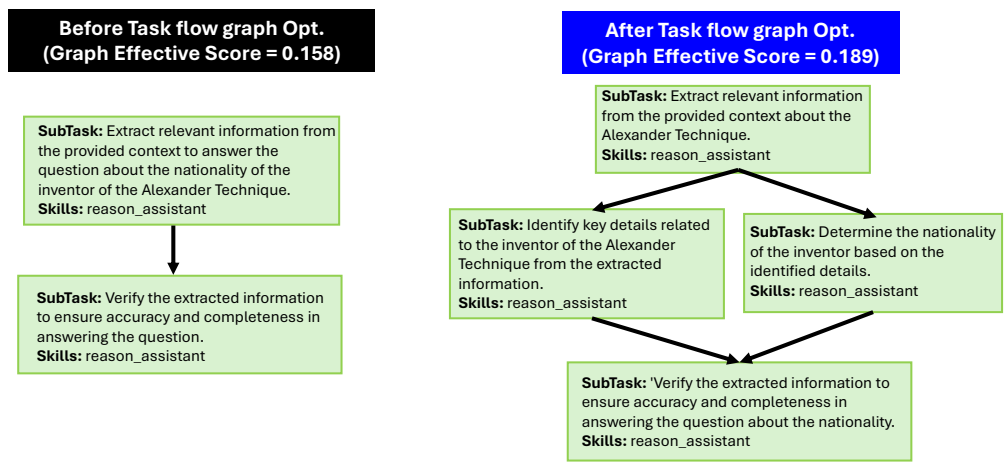

Figure 5: A task flow graph optimization example from HotPotQA.

# Code Represented Workflow Evaluation

In this section, we introduce the workflow evaluation function and its prompt example. Then, we discuss the statistical distribution of workflow evaluation $u(W, K, t)$ between code represented workflow with, and without task flow graph.

## Workflow Evaluation Function

Algorithm 1 presents the workflow evaluation function $u$, which computes multi-objective scores and provides reflections for a code-represented workflow, given the workflow $W$, task flow graph state $K$, and current subtask $t$. The workflow is executed and evaluated by LLM judges using a predefined $num\_trial$, and the evaluation scores are stored in the $eval\_scores$ container (Lines 2–8). The evaluation prompt shown in Figure 6 is used to obtain both the scores and the corresponding reasoning. Next, the average scores for each metric are computed to produce the $combined\_score$, reducing noise from individual LLM judge (Lines 9–10). The reasoning provided by the LLM judges is then leveraged to generate the reflection (Lines 11–12). Finally, the function returns $\bar{C}, \bar{I}, \bar{H}, combined\_score$, and $reflect$ (Line 13).

---

**Algorithm 1:** Workflow Evaluation Function $u(W, k, t)$ Algorithm

---

**Require:** Workflow $W$, Task Flow Graph Status $K$, and Current Subtask $t$
**Ensure:** Multi-Objective Evaluation Scores and Reflection Feedbacks
1: Initialize $eval\_scores \leftarrow$ {Correctness: [], InstructionFollowing: [], MatchHighLevelPlanProgress: [], Reasons: []}
2: **for** $trial \leftarrow 1$ to $num\_trials$ **do**
3:    $e \leftarrow W(t)$    {Execute the workflow $W$}
4:    $scores \leftarrow LLM\_Judge(K, t, e)$    {Use the workflow evaluation prompt with Task=$K + t$, and Output=$e$}
5:    **for** $s$ in $eval\_scores$.keys() **do**
6:      $eval\_scores[s]$.append($scores[s]$)
7:    **end for**
8: **end for**
9: Compute averages: $\bar{C}, \bar{I}, \bar{H}$ for correctness, instruction-following, high-level plan matching from $eval\_scores$
10: $combined\_score \leftarrow 0.4 \cdot \bar{C} + 0.3 \cdot \bar{I} + 0.3 \cdot \bar{H}$
11: $Eval\_feedbacks \leftarrow$ join all $eval\_score[Reasons]$
12: $reflect \leftarrow LLM(Eval\_feedbacks)$
13: **return** $\bar{C}, \bar{I}, \bar{H}, combined\_score, reflect$

---

```
[User Task Description]:
{Task}

[Task Output]:
{Output}

Instruction: Evaluate above Task Output based on the User Task on a scale of 0.0 to 1.0 for the following metrics:
1. InstructionFollowing: How is the Task Output following the Task's instruction?
2. Correctness: How is the correctness of the Task Output to the User Task Request?
3. MatchHighLevelPlanProgress: How is current subtask output match the high-level plan expectation for the subtask?

For each metric, provide a score between 0.0 and 1.0, where 1.0 is best. Please focus on evaluating the task in the [Target Current task request to solve] section.

Return your evaluation as a JSON object with the following format:
{{
  " InstructionFollowing": [score],
  "Correctness": [score],
  "MatchHighLevelPlanProgress": [score],
  "Reasons": "[brief explanation of scores]"
}}
```

Figure 6: The code represented subtask workflow evaluation prompt.

## Code Represented Workflow with/without Task Flow Graph Evaluation Scores Study

We study the differences in workflow evaluation score distributions when applying a task flow graph on top of the code-represented workflow versus using only the code-represented workflow. Figure 7 shows the statistical distribution of workflow evaluations $u(W, K, t)$ for code-represented workflows with and without a task flow graph. These distributions are generated

from 100 sampled questions in the MATH$_{lv5*}$ and MBPP benchmarks. On average, the combined scores of code-represented workflows with a task flow graph are 7% and 11% higher than those without a task flow graph. These observations suggest that task divide-and-conquer via task flow graphs helps generate more stable and correct code-represented workflows.

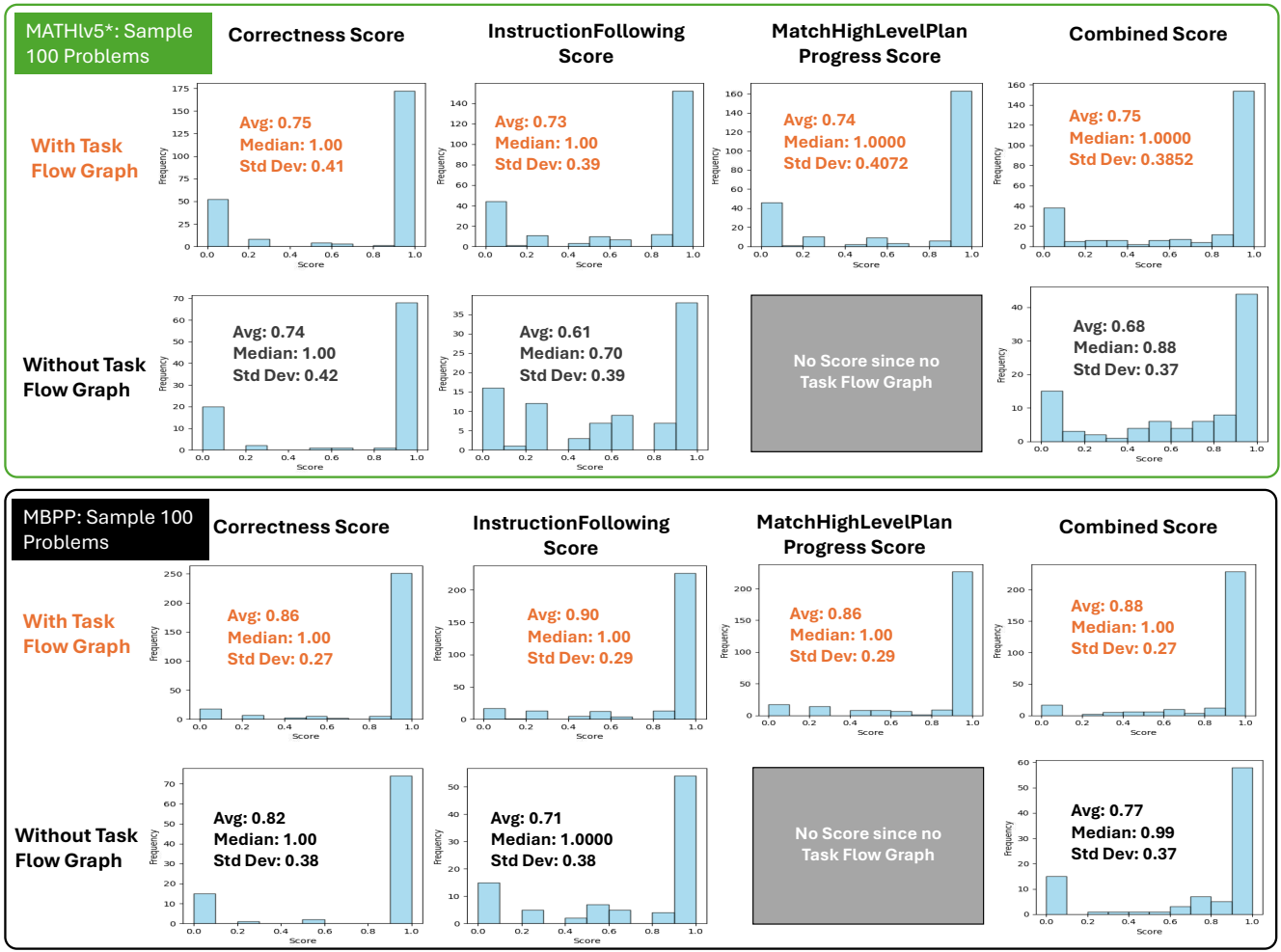

Figure 7: Statistical distribution of workflow evaluation score between code represented workflow with and without task flow graph study on sampled 100 questions in MATH$_{lv5*}$ and MBPP benchmarks.

# Self-Reflection-Guided Evolutionary Algorithm Code Represented Workflow Examples

We present examples of generated code-represented workflows produced by a self-reflection-guided evolutionary algorithm. A total of 35,720 code-represented workflows were generated for various tasks across the Multi-turn QA, Coding, and MATH benchmarks. We use the number of lines in each workflow to indicate its complexity. The minimum, maximum, mean, median, and standard deviation of the workflow complexities are 5.0, 285.0, 64.7, 66.0, and 32.0, respectively.

Figure 8 illustrates the workflow template, where LLM assistants are predefined to generate code-represented workflows, as described in the Preliminary Section. In this template, the workflow function is initially empty and is populated by an LLM, then iteratively optimized through the self-reflection-guided evolutionary algorithm. Figure 9 shows an example of an optimized code-represented workflow. The workflow first employs LLM ensembling to obtain responses from coding assistants and selects the best response among them. Next, a reasoning assistant reflects on the chosen answer to refine it further. Finally, a piece of code is executed without LLM involvement to extract the final numbers and ensure the correctness of the output format.

```python
from assistant_config import get_general_assistant_agent
from advanced_reason_agent_config import get_advanced_reason_agent
from code_agent_config import get_coder_agent
from file_surfer import get_filesurfer_agent


# DONT-Modified-Block-START #
###
# Usable assistants
###
# file_reader: An agent that can read and open a text file given the relative or absolute file path, such as open_local_file, page_up, page_down, find_on_page_ctrl_f,
# find_next, etc. The agent can only read and open a file for each request. Do not ask the agent to show the directory structure.
file_reader, file_reader_description = get_filesurfer_agent()

# general assistant: Assistant who only provide opinions, or summary of information in string text format. For example, code execution result, and file content. The
# info_evaluator assistant can not do any code writing, or traverse directory tasks
general_assistant, general_assistant_description = get_general_assistant_agent()

# coder: A code editing agent with no files loaded (without git version control)
coder, coder_description = get_coder_agent(code_book_manager=None)

# strong_reason_debug_assistant:: Assistant who provide 1) advanced reasoning for debugging and provide insightful comments. 2) directly generate the file that not
# require coding.
strong_reason_debug_assistant, reason_agent_description = get_advanced_reason_agent()

# extract_output function to extract the final response of [coder_assistant, strong_reason_debug_assistant, file_reader, websearch_agent] to string format
# Example 1:
#   response = file_reader.initiate_chat(message=query)
#   response = extract_output(response=response, post_process_func=file_reader_description["post_process_func"])
# Example 2:
#   response = strong_reason_debug_assistant.initiate_chat(message="task")
#   response = extract_output(response=response, post_process_func=reason_agent_description["post_process_func"])
# Example 3:
#   response = coder.initiate_chat(message="task")
#   response = extract_output(response=response, post_process_func=coder_description["post_process_func"])
def extract_output(response, post_process_func=None):
    if post_process_func is not None:
        return post_process_func(response)
    if isinstance(response, str):
        return response

    return response.summary
# DONT-Modified-Block-END #

# EVOLVE-BLOCK-START
def workflow(instruction: str):
    instruction = "### Instruction Background ###\n" + instruction + "\n"

    ... # your designed workflow here.

    return response # response is final answer in str format
# EVOLVE-BLOCK-END
```

Figure 8: Code represented workflow template.

[Problem]: Sam is carrying a 2 gallon bucket of milk to his house from the barn, which is 3 miles away from his house. However, the bucket has a leak in it. For each mile he walks, there is $\frac{2}{3}$ as much milk in the bucket as at the beginning of the mile. How many gallons of milk will be in the bucket when Sam gets home?

[Target Current subtask request to solve]: Calculate the amount of milk remaining after each mile walked. Sam is carrying a 2-gallon bucket of milk, and for each mile walked, there is 2/3 as much milk in the bucket as at the beginning of the mile.

```
# EVOLVE-BLOCK-START
def workflow(instruction: str):

    instruction = "### Instruction Background ###\n" + instruction + "\n"
    # Task 1: Calculate the amount of milk remaining after each mile walked
    current_task = "Task_1_Task_1"
    coder_instruction = instruction + f"\n{current_task}: Calculate and print the amount of milk remaining after each mile walked. Use print(f'After mile {{mile}}: {{amount:.4f}} gallons') for each mile.\n"

    # Generate multiple responses
    num_ensemble = 3
    gen_outs = []
    for i in range(num_ensemble):
        response = coder_assistant.initiate_chat(message=coder_instruction)
        formatted_response = extract_output(response=response, post_process_func=coder_description["post_process_func"])
        gen_outs.append(f"### Generated {i}-th response\n{formatted_response}")

    # Rank responses
    rank_instruction = f"Rank the following code snippets from best to worst based on correctness, efficiency, and adherence to the task:\n\n{''.join(gen_outs)}\n\nOutput the index of the best response."
    rank_response = general_assistant.initiate_chat(message=rank_instruction)
    best_index = int(extract_output(response=rank_response, post_process_func=general_assistant_description["post_process_func"]))

    formatted_response = gen_outs[best_index].split("### Generated")[1].split("-th response\n")[1]

    # Task 2: Calculate the final amount of milk in the bucket after 3 miles
    current_task = "Task_1_Task_2"
    reason_instruction = formatted_response + f"\n{current_task}: Calculate the final amount of milk in the bucket after 3 miles. Include the breakdown of milk amounts after each mile in your response.\n"
    response = strong_reason_debug_assistant.initiate_chat(message=reason_instruction)
    initial_response = extract_output(response=response, post_process_func=reason_agent_description["post_process_func"])

    # Self-reflection
    reflection_prompt = f"""
### Task ###
Provide suggestions, feedback, and reflection on the generated result.

### Task ###
{current_task}: Calculate the final amount of milk in the bucket after 3 miles. Include the breakdown of milk amounts after each mile in your response.

### Previous Generated Result ###
{initial_response}

Reflection:
"""
    reflection_response = strong_reason_debug_assistant.initiate_chat(message=reflection_prompt)
    reflection_output = extract_output(reflection_response, reason_agent_description["post_process_func"])

    # Incorporate reflection into final response
    final_response = f"{initial_response}\n\nReflection and improvements:\n{reflection_output}"

    # Extract the final numerical answer and ensure it's in the required boxed format
    import re

    numerical_answer = re.search(r'\d+\.\d+', final_response)
    if numerical_answer:
        boxed_answer = f"$\\boxed{{{numerical_answer.group()}}}$"
        final_response = f"{final_response}\n\nFinal Answer: {boxed_answer}"
    else:
        final_response = f"{final_response}\n\nError: No numerical answer found."

    return final_response
# EVOLVE-BLOCK-END
```

Figure 9: An example of generated code-represented workflows from self-reflection-guided evolutionary algorithm that includes ensembling approach, ranking or voting, reflection on top of coding and reasoning.



\end{document}